\pdfoutput=1
\documentclass[11pt]{article}

\usepackage[preprint]{acl}
\usepackage{times}
\usepackage{latexsym}
\usepackage[T1]{fontenc}
\usepackage[utf8]{inputenc}
\usepackage{microtype}
\usepackage{inconsolata}
\usepackage{comment}
\usepackage{graphicx}
\usepackage{amsmath}
\usepackage{amssymb}
\usepackage{amsfonts}
\usepackage{array}
\usepackage{multirow}
\usepackage{booktabs}
\usepackage{adjustbox}
\usepackage{subcaption}
\usepackage{stmaryrd}
\usepackage{siunitx}
\usepackage{caption}
\usepackage{url}
\usepackage{bm}
\usepackage{longtable}
\usepackage{algorithm,algpseudocode}

\DeclareMathOperator{\KL}{KL}

\DeclareMathOperator*{\argmin}{arg\,min}

\newcommand\blfootnote[1]{
  \begingroup
  \renewcommand\thefootnote{}\footnote{#1}
  \addtocounter{footnote}{-1}
  \endgroup
}

\title{Domain Mixture Design via Log-Likelihood Differences\\for Aligning Language Models with a Target Model}

\author{
{\bf Ryo Kishino}${}^{1}$\qquad 
{\bf Riku Shiomi}${}^{1}$\qquad {\bf Hiroaki Yamagiwa}${}^{2}$\thanks{Work done while at Kyoto University.}\\ 
{\bf Momose Oyama}${}^{1,3}$ \qquad {\bf Hidetoshi Shimodaira}${}^{1,3}$\\
${}^{1}$Kyoto University\quad
${}^{2}$SB Intuitions\quad
${}^{3}$RIKEN\\
\texttt{\{kishino.ryo.32s, shiomi.riku.52p\}@st.kyoto-u.ac.jp},\\
\texttt{hiroaki.yamagiwa@sbintuitions.co.jp,}\,
\texttt{oyama.momose@sys.i.kyoto-u.ac.jp},\\
\texttt{shimo@i.kyoto-u.ac.jp}
}

\begin{document}
\maketitle
\begin{abstract}
Instead of directly distilling a language model, this study addresses the problem of aligning a base model with a target model in distribution by designing the domain mixture of training data for pretraining or continued pretraining as a fixed training recipe. We propose a method for determining domain weights by viewing models as points in log-likelihood space and aligning the training update direction with the direction toward the target model. Experiments with NanoGPT show that the proposed method consistently reduces the KL divergence between the trained base model and the target model relative to training with Pile-original weighting. Although knowledge distillation remains more effective when available, the proposed method achieves meaningful alignment, and downstream task performance also tends to become closer to that of the target model.
\blfootnote{Our code is available at \url{https://github.com/shimo-lab/modelmap}.}
\end{abstract}

\section{Introduction}\label{sec:intro}

The properties of large language models depend not only on their architecture and model size but also on the distribution of text used for training~\cite{albalak2024a}. In particular, large-scale corpora such as the Pile~\cite{gao2020pile800gbdatasetdiverse} consist of multiple domains, and the mixture weights over these domains strongly affect model behavior~\cite{NEURIPS2023_dcba6be9}. Therefore, designing the domain mixture of the training data is an important problem for enabling models to acquire desired properties.

In this study, we address the problem of determining the training text distribution for updating a base model so as to align it with a target model in distribution. A common approach is to directly imitate the target model through knowledge distillation~\cite{hinton2015distillingknowledgeneuralnetwork} or by training on text generated by the target model~\cite{kim-rush-2016-sequence}. However, such methods require access to the teacher model and incur additional generation or inference costs. Moreover, even when the domain weights used in prior training are known, simply reusing them is not necessarily optimal in the current setting. We therefore aim to achieve model alignment by designing the domain weights of the training text as a \emph{fixed training recipe} specified before training begins.
The estimated domain weights can serve not only as a target-informed starting point for further training under different settings, but also as an interpretable summary of the domain-level shift from the base model toward the target.

\begin{figure}[t]
  \centering
  \includegraphics[width=1.0\linewidth]{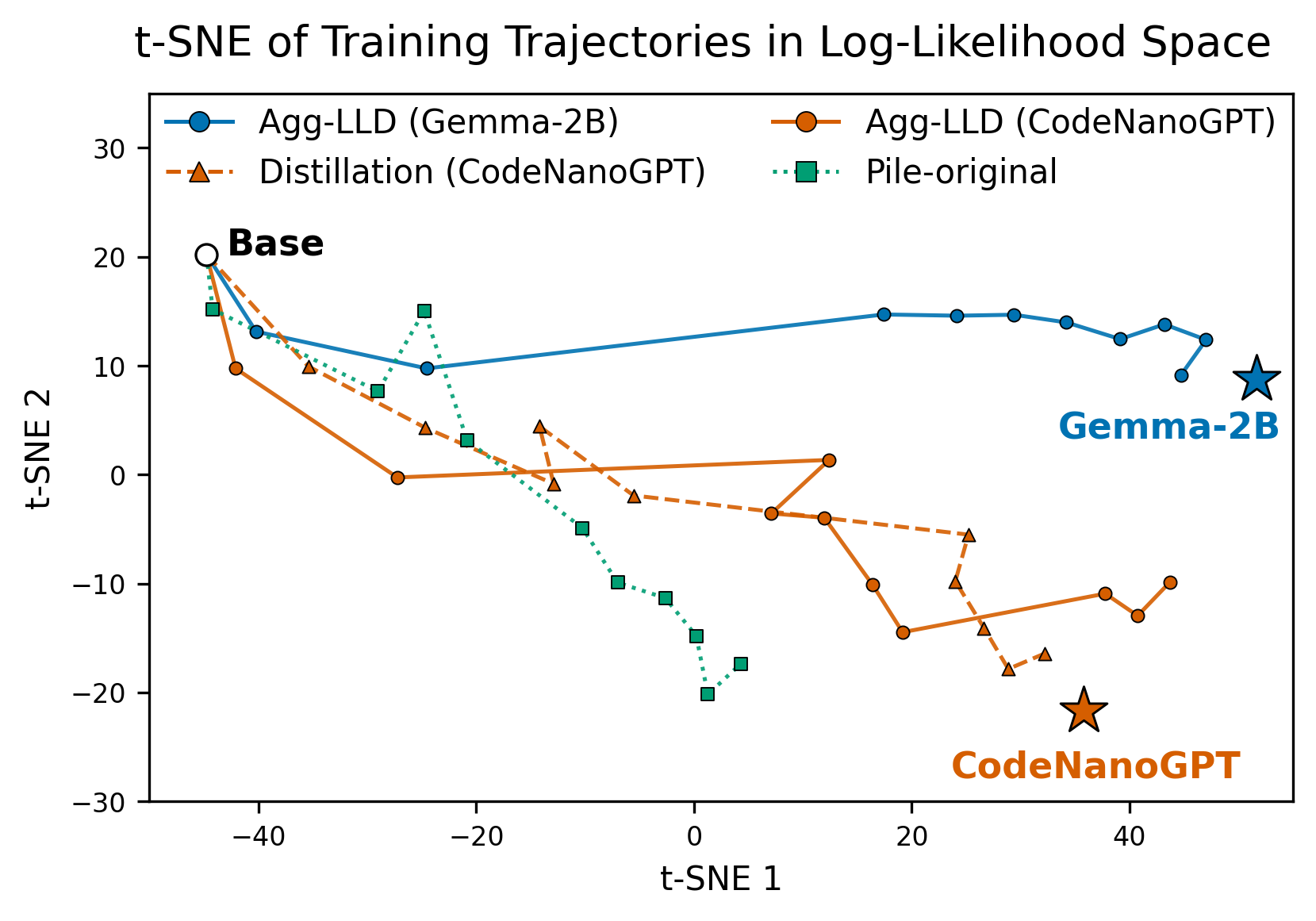}
  \caption{
t-SNE visualization of training trajectories in the log-likelihood vector space during continued pretraining from pretrained NanoGPT.
We consider Gemma-2B and CodeNanoGPT as target models.
For each target, we show the trajectory obtained using the fixed domain weights estimated by the proposed method (aggregated-LLD), along with a reference trajectory obtained using Pile-original weights.
For CodeNanoGPT, we also show the trajectory obtained by knowledge distillation.
Each point represents a checkpoint taken every 1k training steps, and the stars denote the target models. Compared with Pile-original weighting, the proposed method yields trajectories that move more consistently toward the target models and, for CodeNanoGPT, produces a trajectory qualitatively similar to that of distillation.
See Sections~\ref{subsec:ft-kl-target} and \ref{subsec:distill-comparision} for details.
  }
  \label{fig:traj-original-proposal}
\end{figure}

We represent a model as a vector of mean log-likelihoods over multiple text collections~\cite{oyama-etal-2025-mapping}. In this study, we use the domains of the Pile~\cite{gao2020pile800gbdatasetdiverse} as these text collections. Directly comparing different models in parameter space is difficult because of architectural mismatches and permutation symmetries of network weights induced by hidden-unit reordering~\cite{HECHTNIELSEN1990129}. In contrast, mean log-likelihood vectors computed on a set of text collections provide a common basis for comparison across models. With this representation, model alignment can be viewed as a geometric problem of moving the base model toward the target model, leading to a rule for determining mixture proportions based on the domain-level log-likelihood differences (LLD) between the target and base models.

In experiments using NanoGPT as the base model, determining the domain weights for the Pile using the proposed method consistently reduces the KL divergence between the trained NanoGPT and the target models, Gemma-2B or CodeGemma-2B, compared with training using Pile-original weights.
Moreover, for target models that share the same tokenizer, the proposed method provides a useful alternative to knowledge distillation.
Fig.~\ref{fig:traj-original-proposal} visualizes how the training trajectories obtained by the proposed method move toward the corresponding target models in log-likelihood space, compared with those obtained using Pile-original weighting or knowledge distillation.

\section{Background and Related Work}\label{sec:related_work}

\subsection{Data selection and mixing for LLMs}

\begin{figure}[t]
  \centering
  \includegraphics[width=1.0\linewidth]{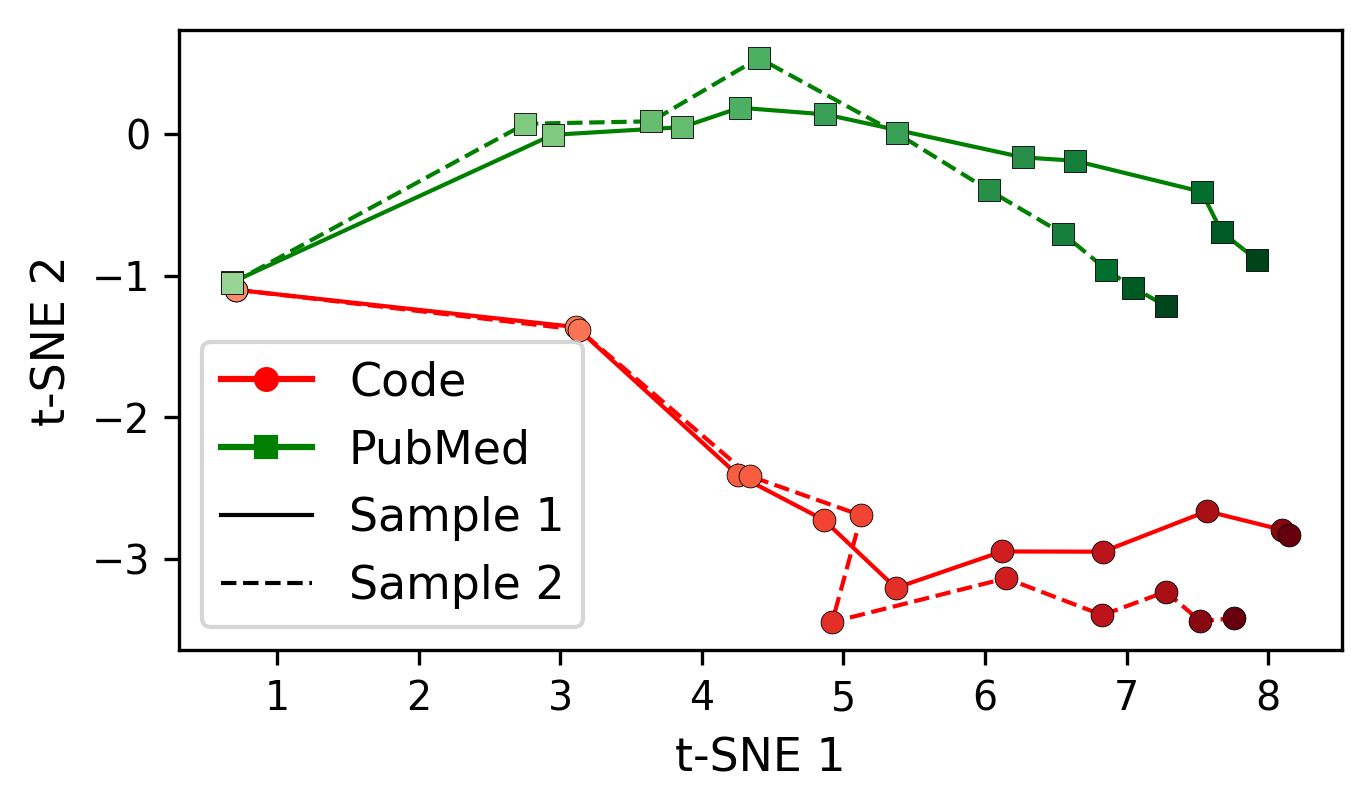}
  \caption{
t-SNE visualization of NanoGPT training trajectories for two training datasets independently sampled from each of two domain mixtures over the Pile. Each point represents a checkpoint taken every 1k training steps, and color intensity indicates the number of training steps. Differences in trajectory are much larger across domain mixtures than across datasets sampled from the same mixture. The two mixtures are obtained by quadrupling the weights of either the code-related domains or the PubMed-related domains relative to the original Pile weights. See Appendix~\ref{app:experimental-setup-detail} for details.
}
  \label{fig:domain-effect}
\end{figure}

Selecting and mixing training data has become an important issue in LLM training, as performance depends not only on the amount of data but also on its distribution~\cite{albalak2024a}. In particular, for multi-domain corpora such as the Pile~\cite{gao2020pile800gbdatasetdiverse}, the choice of domain weights substantially affects model properties and can lead to markedly different training trajectories in log-likelihood space (Fig.~\ref{fig:domain-effect}). This has led to active research on optimizing training data selection and domain weights for downstream tasks, domain adaptation, and training efficiency~\cite{NEURIPS2023_ac662d74, NEURIPS2023_dcba6be9, ye2025limo, NEURIPS2024_b206d54f, dohmatob2026why, chen2026olmixframeworkdatamixing}.

As a classical data selection method, \citet{moore-lewis-2010-intelligent} proposed ranking texts in a general-domain corpus based on the log-likelihood difference between a language model trained on the target domain and another trained on the general domain. Their method and ours both use differences in model likelihoods as a signal for training-data design. However, \citet{moore-lewis-2010-intelligent} use this signal to rank individual texts for domain adaptation, whereas our method aggregates log-likelihood differences at the domain level and uses them to determine mixture weights for moving a base model toward a specified target model in log-likelihood space.

Recently, several studies have explored how to choose effective domain weights for language model training. \citet{NEURIPS2023_dcba6be9} proposed DoReMi, trained a small proxy model with group distributionally robust optimization using domain-wise excess losses relative to a reference model, and used the resulting domain weights to resample data for training a larger model.
\citet{NEURIPS2024_b206d54f}, in a knowledge distillation setting, dynamically updated the domain weights of the distillation data according to the domain-level perplexity ratios between the teacher and the student. In contrast, our study aims to align a base model with a specific target model by designing domain weights as a fixed training recipe, without using distillation. Our study can therefore be viewed as distillation-free data mixture design for distributional alignment to a target model.

\subsection{Log-likelihood vector}

As a framework for comparing models in a shared space, \citet{oyama-etal-2025-mapping} introduced, for a language model $p$, the text-level log-likelihood vector over a set of $N$ texts $\mathcal{D}=\{x_1,\ldots,x_N\}$,
\begin{equation}\label{eq:log-likelihood-vector}
\bm \ell^{\text{text}} = (\log p(x_1), \ldots, \log p(x_N))^\top \in \mathbb{R}^N,
\end{equation}
which serves as a representation of $p$. For $M$ models (or checkpoints along a training trajectory) $p_1,\ldots,p_M$, let
\[
\bm L = (\bm \ell^{\text{text}}_1,\ldots,\bm \ell^{\text{text}}_M)^\top \in \mathbb{R}^{M\times N}
\]
and define
\[
\bm Q = (\bm q_1,\ldots,\bm q_M)^\top \in \mathbb{R}^{M\times N}
\]
as the matrix obtained by double-centering\footnote{Centering the log-likelihood matrix row-wise over texts and column-wise over models.} $\bm L$. Then, the KL divergence between $p_i$ and $p_j$ is approximated as
\begin{equation}\label{eq:KL}
2\KL(p_i,p_j) \approx \|\bm q_i-\bm q_j\|^2 / N.
\end{equation}
This approximation enables comparison of models with different architectures in a shared geometric space. Moreover, log-likelihood vectors have been shown to be useful for predicting downstream task performance.
The computational cost of constructing such model maps can also be reduced by resampling texts according to the variance of their log-likelihoods across models~\cite{oyama-etal-2025-likelihood}.
Log-likelihood vectors have further been applied to the analysis of training trajectories during pretraining~\cite{kishino-etal-2026-establishing} and to the analysis of how training data contributes to learning efficiency~\cite{harada-etal-2025-massive, aoki2025learningdomain}.

Building on this framework, we use the text-level log-likelihood vector to analyze training trajectories and the domain-level log-likelihood vector, consisting of mean log-likelihoods for each domain, to design domain weights for training.

\section{Method}\label{sec:method}

In this study, we consider domain weights for distributionally aligning a base model with a target model in log-likelihood space. Our main objective is to design a single set of domain weights fixed once before training. We first formulate the problem and derive a theoretical criterion that determines domain weights iteratively based on update-direction alignment as a first pass. We then aggregate the resulting weights into a form that is easier to use as a fixed recipe in a second pass.

\subsection{Text distribution}

For each domain \(k \in \{1,\ldots,K\}\), let \(r_k(x)\) denote the domain-specific text distribution\footnote{For simplicity, we refer to the \(k\)-th domain simply as domain \(k\). In practice, the domains include, for example, GitHub and arXiv.}. Let
\[
\bm\pi=(\pi_1,\ldots,\pi_K)^\top \in \Delta^{K-1}
\]
denote the domain weights, where
\[
\Delta^{K-1}=\left\{\bm\pi:\sum_{k=1}^K \pi_k=1,\ \pi_k\ge0\right\}
\]
is the \((K-1)\)-dimensional probability simplex. We then define the corresponding mixture text distribution as
\[
r_{\bm\pi}(x)=\sum_{k=1}^K \pi_k r_k(x).
\]

\subsection{Domain-level log-likelihood vector}

For each \(k\), let \(\mathcal D_k\) denote a set of texts sampled from \(r_k\), and let
\[
\mathcal D = \bigcup_{k=1}^K \mathcal D_k
\]
be the union of these sets. We refer to \(\mathcal D\) as the evaluation corpus.

We consider a base model \(p_{\bm\theta}(x)\) and a target model \(p_{\text{tgt}}(x)\), where \(\bm\theta\in\mathbb{R}^P\) denotes the parameters of the base model. The goal of this study is to align the base model with the target model through pretraining or continued pretraining. The two models need not share the same architecture.

For the base model \(p_{\bm\theta}(x)\), we define the domain-level log-likelihood vector on the evaluation corpus \(\mathcal D\) by
\begin{equation}\label{eq:domain-log-likelihood-vector}
{\bm\ell}(\bm\theta)=
\begin{pmatrix}
\frac{1}{|\mathcal{D}_1|}\sum_{x\in \mathcal D_1}\log p_{\bm\theta}(x)\\
\vdots\\
\frac{1}{|\mathcal{D}_K|}\sum_{x\in \mathcal D_K}\log p_{\bm\theta}(x)
\end{pmatrix}
\in\mathbb{R}^K.
\end{equation}
Similarly, let \(\bm\ell_{\text{tgt}}\in\mathbb{R}^K\) denote the domain-level log-likelihood vector for the target model \(p_{\text{tgt}}(x)\).

\subsection{Derivation of a rule for determining domain weights}
\begin{figure}[t]
    \centering
    \includegraphics[width=1.0\linewidth]{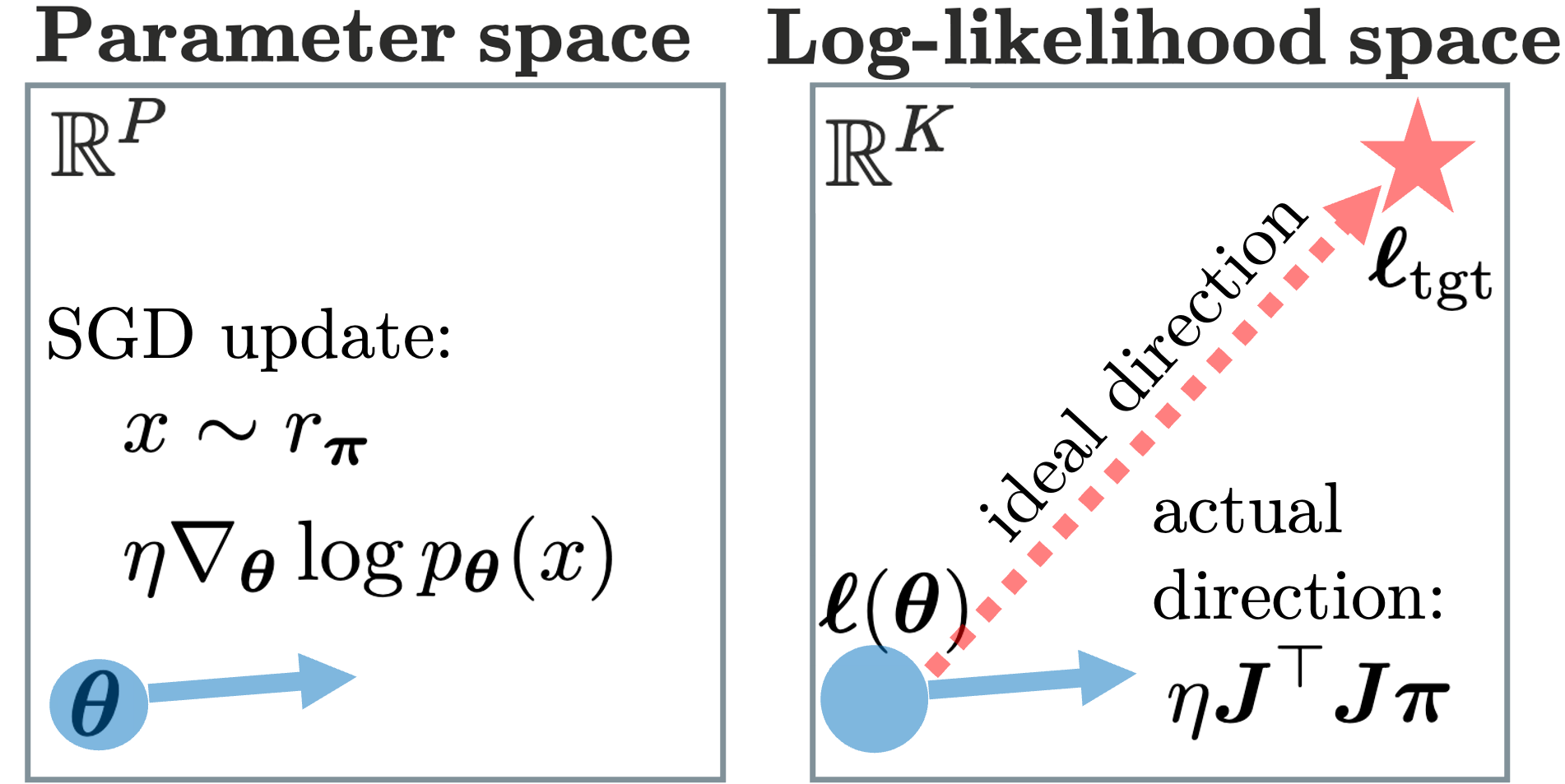}
    \caption{
    Illustration of the update direction in log-likelihood space when the base model \(p_{\bm\theta}\) is updated along the gradient direction in parameter space using a text sampled from the mixture distribution \(r_{\bm\pi}\). Our goal is to estimate \(\bm\pi\) so that the blue update direction aligns with the red direction toward the target model.
    }
    \label{fig:weight_logp_illust}
\end{figure}

We first derive how training changes the log-likelihood vector. Assuming SGD with learning rate \(\eta>0\), the parameter update with a text \(x\sim r_{\bm \pi}\) is given by
\[
\Delta \bm\theta := \eta\nabla_{\bm\theta}\log p_{\bm\theta}(x),\quad \bm\theta\leftarrow \bm\theta + \Delta\bm\theta.
\]
Let the domain-level Jacobian on the evaluation corpus be
\[
{\bm J}(\bm\theta) = \nabla_{\bm\theta}{\bm\ell}(\bm \theta)^\top\in\mathbb{R}^{P\times K}.
\]
Then, the expected change in the log-likelihood vector is approximated as
\begin{equation}\label{eq:sgd-logp-diff}
    \mathbb{E}_{x\sim r_{\bm \pi}} [{\bm\ell}(\bm\theta+\Delta\bm\theta)-{\bm\ell}(\bm\theta)] \approx {\eta}{\bm J}(\bm\theta)^\top {\bm J}(\bm\theta) \bm \pi.
\end{equation}
See Appendix~\ref{app:proof-sgd-logp-diff} for the proof.

For simplicity, we omit the explicit dependence on $\bm\theta$ and write $\bm\ell$ and $\bm J$. We refer to $\bm J^\top \bm J$ as the Gram matrix\footnote{$\bm J$ may be viewed as a score-like matrix over domains, and $\bm J \bm J^\top$ corresponds to the empirical Fisher information matrix in parameter space.}. Equation~\eqref{eq:sgd-logp-diff} implies that the expected update direction in log-likelihood space is approximated by $\eta{\bm J}^\top {\bm J}\bm\pi$, shown as the blue arrow in Fig.~\ref{fig:weight_logp_illust}. Our goal is to align this direction with the target direction ${\bm \ell}_{\text{tgt}} - {\bm \ell}$, shown as the red arrow.

If we ignore the constraint \(\bm\pi \in \Delta^{K-1}\), we seek \(\bm\pi\) such that the expected update direction matches the target direction\footnote{Since only the direction matters, we omit \(\eta\).}:
\[
{\bm J}^\top {\bm J}\bm \pi = {\bm\ell}_{\text{tgt}} - {\bm\ell}.
\]
The unconstrained weight vector satisfying this relation is
\[
\tilde{\bm\pi} = ({\bm J}^\top {\bm J})^{-1}({\bm\ell}_{\text{tgt}} - {\bm\ell}).
\]
To obtain domain weights in \(\Delta^{K-1}\), we consider the optimization problem
\begin{equation}\label{eq:reg-problem}
    \max_{\bm\pi\in\Delta^{K-1}}~\bm \pi^\top \tilde{\bm\pi} - \tau \KL(\bm\pi,\bm\pi').
\end{equation}
Here, \(\tau>0\) is the temperature parameter controlling the strength of regularization, and \(\KL(\bm\pi,\bm\pi')\) is a regularization term that encourages domain weights close to the reference weights $\bm\pi'$, thereby avoiding extreme weights
\footnote{
Hence, \eqref{eq:reg-problem} can be interpreted equivalently as a one-step update of mirror descent~\cite{nemirovskij1983problem} with negative entropy as the distance-generating function (Appendix~\ref{app:mirror-descent}), or as the multiplicative weights update method~\cite{v008a006}.}.
The solution to \eqref{eq:reg-problem} is the softmax of \(\tilde{\bm\pi}/\tau\):
\begin{equation}\label{eq:opt-domain-ratio}
    \pi_k = \frac{\pi'_k\exp(\tilde \pi_k / \tau)}{\sum_{k'=1}^K  \pi'_{k'}\exp(\tilde \pi_{k'}/\tau) }.
\end{equation}
The proof is given in Appendix~\ref{app:proof-opt-domain-ratio}.

In practice, computing the Gram matrix \(\bm{J}^\top \bm J\) requires \(O(K^2 P)\) time, which imposes substantial computational and memory costs for large language models with a huge number of parameters \(P\). We therefore omit \(({\bm J}^\top {\bm J})^{-1}\), which provides sensitivity adjustment across domains, and instead use a heuristic approximation given by the softmax of \(({\bm\ell}_{\text{tgt}} - {\bm\ell})/\tau\):
\begin{equation}\label{eq:raw-domain-ratio}
    \pi_k = \frac{\pi'_k\exp(({\ell}_{\mathrm{tgt},k} - {\ell}_k)/\tau)}{\sum_{k'=1}^K \pi'_{k'}\exp(({\ell}_{\mathrm{tgt},k'} - {\ell}_{k'})/\tau)}.
\end{equation}
In what follows, we use this approximation as a practical mixture design method, while treating the method with the adjustment matrix \(({\bm J}^\top {\bm J})^{-1}\) as a theoretical reference method.
 
\subsection{Iterative estimation of domain weights and aggregation}

As \eqref{eq:raw-domain-ratio} indicates, the domain weights depend on \({\bm\ell}(\bm\theta)\), so the theoretically desirable weights change during training. We therefore adopt a two-pass procedure. In the first pass, the domain weights are computed at selected update steps and then aggregated into a single set of weights, \(\bm\pi^*\), to be used as a fixed training recipe. In the second pass, the model is trained using \(\bm\pi^*\), without further reference to the target model.

Here, \(\mathcal D\) denotes the evaluation corpus used to compute the domain-level log-likelihood vector, and \(\mathcal C_k\) denotes the training corpus for domain \(k\). Let
\[
\mathcal T = \{t_1, \ldots, t_{|\mathcal T|}\}
\]
denote the set of training steps at which domain weights are estimated. Given domain weights \(\bm\pi^{(t)}\) at step \(t\), we define the fixed domain weights \(\bm\pi^*\) by geometric-mean aggregation over \(t\in\mathcal T\):
\begin{align}\label{eq:domain-ratio-aggregate}
    \pi_k^* \propto \left(\prod_{t\in\mathcal T} \pi_k^{(t)}\right)^{1/|\mathcal T|}.
\end{align}
This coincides with the point in \(\Delta^{K-1}\) that minimizes the sum of KL divergences\footnote{When the KL divergence is replaced by the squared Euclidean distance, the solution to \(\min_{\bm\pi\in\Delta^{K-1}}\sum_{t\in\mathcal T}\|\bm\pi - \bm\pi^{(t)}\|^2\) is \(\bm\pi^* = \frac{1}{|\mathcal T|}\sum_{t\in\mathcal T}\bm\pi^{(t)}\). See Appendix~\ref{app:aggregate} for the difference between arithmetic-mean aggregation and geometric-mean aggregation.} to \(\bm\pi^{(t)}\):
\begin{align}\label{eq:domain-ratio-aggregate-optimization}
   \bm\pi^* =  \argmin_{\bm\pi \in \Delta^{K-1}}\sum_{t\in\mathcal T} \KL(\bm\pi,\bm\pi^{(t)}).
\end{align}
See Appendix~\ref{app:proof-domain-ratio-aggregate} for the proof.

We call adjusted-LLD the theoretical reference method that updates domain weights at each step in \(\mathcal T\) according to \eqref{eq:opt-domain-ratio}, using the sensitivity adjustment \((\bm J^\top \bm J)^{-1}\). In contrast, we call iterative-LLD the method that updates domain weights at each step according to \eqref{eq:raw-domain-ratio} without the sensitivity adjustment. 
We call aggregated-LLD the two-pass method that aggregates the iterative-LLD weights over the steps in \(\mathcal T\) via \eqref{eq:domain-ratio-aggregate} to estimate fixed domain weights \(\bm\pi^*\), and then trains the base model using them. Algorithm~\ref{algo:domain-ratio} summarizes the first pass.

The practical motivation for this two-pass design is that the final domain mixture can be fixed and disclosed before the second-pass training begins. This provides a predetermined and reproducible training recipe, allows the training data and sampling pipeline to be prepared in advance, and facilitates auditing and comparison under a fixed data budget.

\begin{algorithm}[t]
\caption{First-pass estimation of the fixed domain weights $\bm{\pi}^*$ for aggregated-LLD}
\label{algo:domain-ratio}
\begin{algorithmic}[1]
\Require base model $p_{\bm\theta}$, target model $p_{\text{tgt}}$, evaluation corpus $\mathcal{D}$, domain-specific training corpora $\{\mathcal{C}_k\}_{k=1}^K$, update steps $\mathcal{T}$, total training steps $t_{\text{max}}$, temperature $\tau$
\Ensure domain weights $\bm\pi^*$
\Statex
\State Compute the domain-level log-likelihood vector ${\bm\ell}_{\text{tgt}}$ using \eqref{eq:domain-log-likelihood-vector}
\For{$t = 0,\ldots,t_{\text{max}}$}
    \If{$t\in\mathcal T$ or $t=0$}
        \State Compute ${\bm\ell}(\bm\theta)$ using \eqref{eq:domain-log-likelihood-vector}
        \State Estimate $\bm\pi^{(t)}$ using \eqref{eq:raw-domain-ratio}
    \Else
        \State Set $\bm\pi^{(t)}\leftarrow \bm\pi^{(t-1)}$
    \EndIf
    \State Sample a domain $k$ according to the domain weights $\bm\pi^{(t)}$
    \State Update $\bm\theta$ using a batch sampled from $\mathcal{C}_k$
\EndFor
\State Aggregate $\{\bm\pi^{(t)}\}_{t\in\mathcal T}$ using \eqref{eq:domain-ratio-aggregate} to obtain $\bm\pi^*$
\end{algorithmic}
\end{algorithm}

\section{Experiments}\label{sec:experiments}

In this section, we evaluate how well alignment with the target model can be achieved by training the base model using domain weights designed by the proposed method.
We examine four aspects: (i) the validity of domain-weight estimation for target models with known domain weights, (ii) alignment with publicly available LLMs, (iii) comparison with knowledge distillation, and (iv) similarity in downstream task performance. Details of the experimental setup are provided in Appendix~\ref{app:experimental-setup-detail}.

\subsection{Experimental setup}\label{subsec:experimental-setup}

\paragraph{Datasets.}
Both the evaluation and training texts are taken from the Pile Uncopyrighted\footnote{\url{https://huggingface.co/datasets/monology/pile-uncopyrighted}}, which contains \(K=17\) domains\footnote{Five copyrighted domains have been removed from the original 22 domains in the Pile.}. For the evaluation corpus \(\mathcal{D}\) used for computing log-likelihood vectors, we use the same 10,000 texts as in \citet{oyama-etal-2025-mapping}\footnote{\url{https://github.com/shimo-lab/modelmap}}. These texts are chunked into 1024-byte segments, and the approximate KL divergence is measured on this text set in bits per byte using \eqref{eq:KL}. Unless otherwise specified, we use ``KL divergence'' in the experimental results to refer to this approximation.
By contrast, each domain corpus \(\mathcal{C}_k\) used to train the base model is also derived from the Pile Uncopyrighted, but consists of original texts tokenized and stored separately by domain. During training, each batch is constructed from a domain sampled according to the domain weights. The total number of tokens per training step is \(2^{19}\approx 500\mathrm{k}\).

\paragraph{Base/target model settings.}
In our experiments, we use NanoGPT~\cite{Karpathy2022} with 124M parameters as the base model. To examine whether the proposed method remains effective even when the base model has already acquired general knowledge, we consider not only a randomly initialized NanoGPT but also a NanoGPT pretrained on FineWeb 100BT~\cite{NEURIPS2024_370df50c}. As target models, we use publicly available models such as Gemma-2B~\cite{gemmateam2024gemmaopenmodelsbased} and CodeGemma-2B~\cite{codegemmateam2024codegemmaopencodemodels}, as well as NanoGPT pretrained on the Pile with controlled domain weights. In particular, we define CodeNanoGPT as a model trained for 100k steps on the Pile with the weights of GitHub, Stack Exchange, Ubuntu IRC, and Hacker News quadrupled relative to their original values, increasing the total proportion of code-related domains from 18.9\% to 75.6\%. See Appendix~\ref{app:experimental-setup-detail} for the hyperparameters used in these pretraining runs.

\paragraph{Computation of the domain-level log-likelihood vector.}
The domain-specific corpus \(\mathcal{C}_k\) used for training and the evaluation text set \(\mathcal{D}\) differ in how texts are segmented. The text sequences used to construct training batches have a fixed token length, whereas the evaluation texts vary in token length, as shown in Table~\ref{tab:the-pile-domain-numtexts} in Appendix~\ref{app:experimental-setup-detail}. To make log-likelihoods from the two settings comparable, we normalize the domain-level mean log-likelihoods in \eqref{eq:domain-log-likelihood-vector} by the mean token length of each domain\footnote{Equivalently, this is the same as dividing the total log-likelihood by the total number of tokens for each domain.}.

\begin{figure*}[t]
  \centering
  \includegraphics[width=1.0\linewidth]{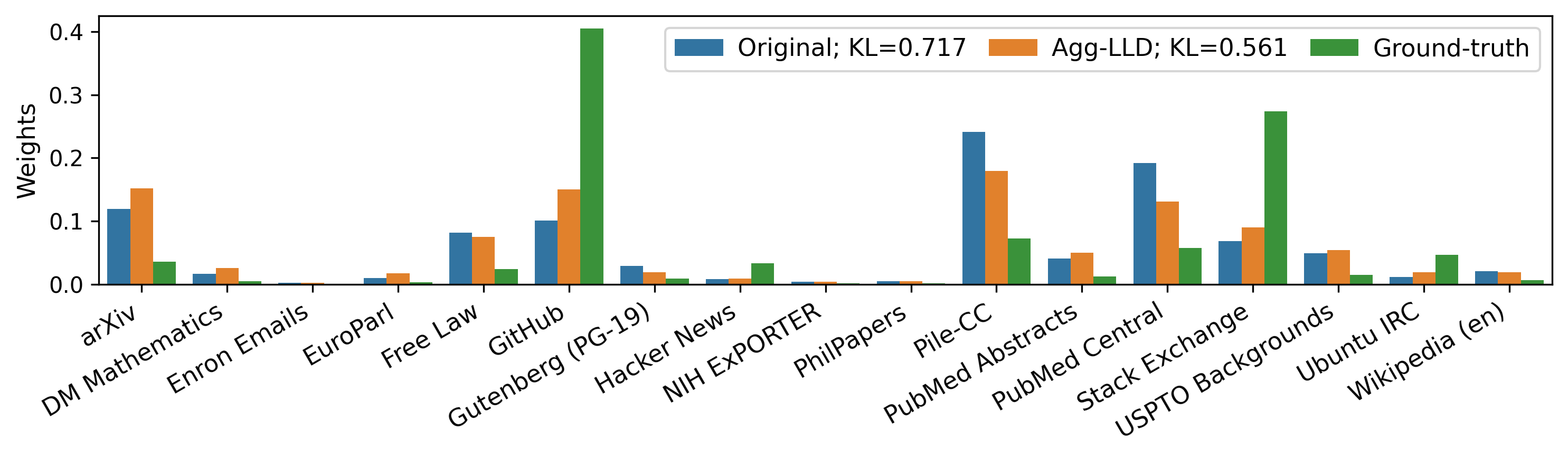}
  \caption{
Comparison of domain weights estimated by aggregated-LLD and Pile-original weights against the ground-truth. The target model is CodeNanoGPT, pretrained with known domain weights, and the base model is a randomly initialized NanoGPT. The values in the legend denote the KL divergence from the ground-truth weights.}
  \label{fig:visualize-domain-ratio}
\end{figure*}

\paragraph{Base model training.}
The base model is trained for 10k steps in both pretraining and continued pretraining.
At each step, one domain is sampled according to the domain weights, and the batch is constructed from that domain. The domain weights are updated iteratively at steps 0, 1, 2, 4, 8, 16, 32, 64, 128, 256, 512, 1k, 2k, \ldots, 9k. This follows the Pythia setting~\cite{3618408.3618510}, in which checkpoints are saved densely during the warmup. For iterative-LLD and adjusted-LLD, training proceeds while iteratively estimating domain weights at these steps using the current base model. In contrast, for aggregated-LLD, we first aggregate the domain weights obtained by iterative-LLD to determine a fixed set of domain weights, and then train the base model using these fixed domain weights. The softmax temperature in \eqref{eq:raw-domain-ratio} for domain-weight estimation is set to $\tau=1$; ablations are in Appendix~\ref{app:tau-ablation}.
We use the original domain weights of the Pile as the reference weights \(\bm\pi'\). The specific values are listed in Table~\ref{tab:the-pile-domain-ratio} in Appendix~\ref{app:experimental-setup-detail}.
We use AdamW~\cite{DBLP:conf/iclr/LoshchilovH19} as the optimizer in all experiments.

\subsection{Comparison with ground-truth weights}\label{subsec:experiment-visualize-domain-ratio}

To examine how aggregated-LLD estimates domain weights, we consider a setting in which the target model is CodeNanoGPT, whose ground-truth weights are known, and the base model is a randomly initialized NanoGPT. Fig.~\ref{fig:visualize-domain-ratio} shows the domain weights \(\bm\pi^*\) estimated by Algorithm~\ref{algo:domain-ratio}.

The KL divergence from the ground-truth domain weights is 0.717 for Pile-original weights and 0.561 for the estimated weights. This indicates that the proposed method yields domain weights closer to the ground-truth than Pile-original weights. In particular, for code-related domains whose weights are increased in the ground-truth, such as GitHub and Stack Exchange, the estimated weights are broadly consistent with the ground-truth trend. Although the estimated weights do not exactly match the ground-truth, they capture the target model's overall domain weighting pattern.

\subsection{Distributional alignment via estimated domain weights}\label{subsec:ft-kl-target}

We evaluate how much the KL divergence between the base model and the target model is reduced when the base model is trained with the estimated domain weights. We compare Pile-original domain weights, adjusted-LLD, iterative-LLD, and aggregated-LLD.

\begin{figure}[t]
    \centering
    \begin{subfigure}[t]{1.0\linewidth}
        \centering
        \includegraphics[width=1.0\linewidth]{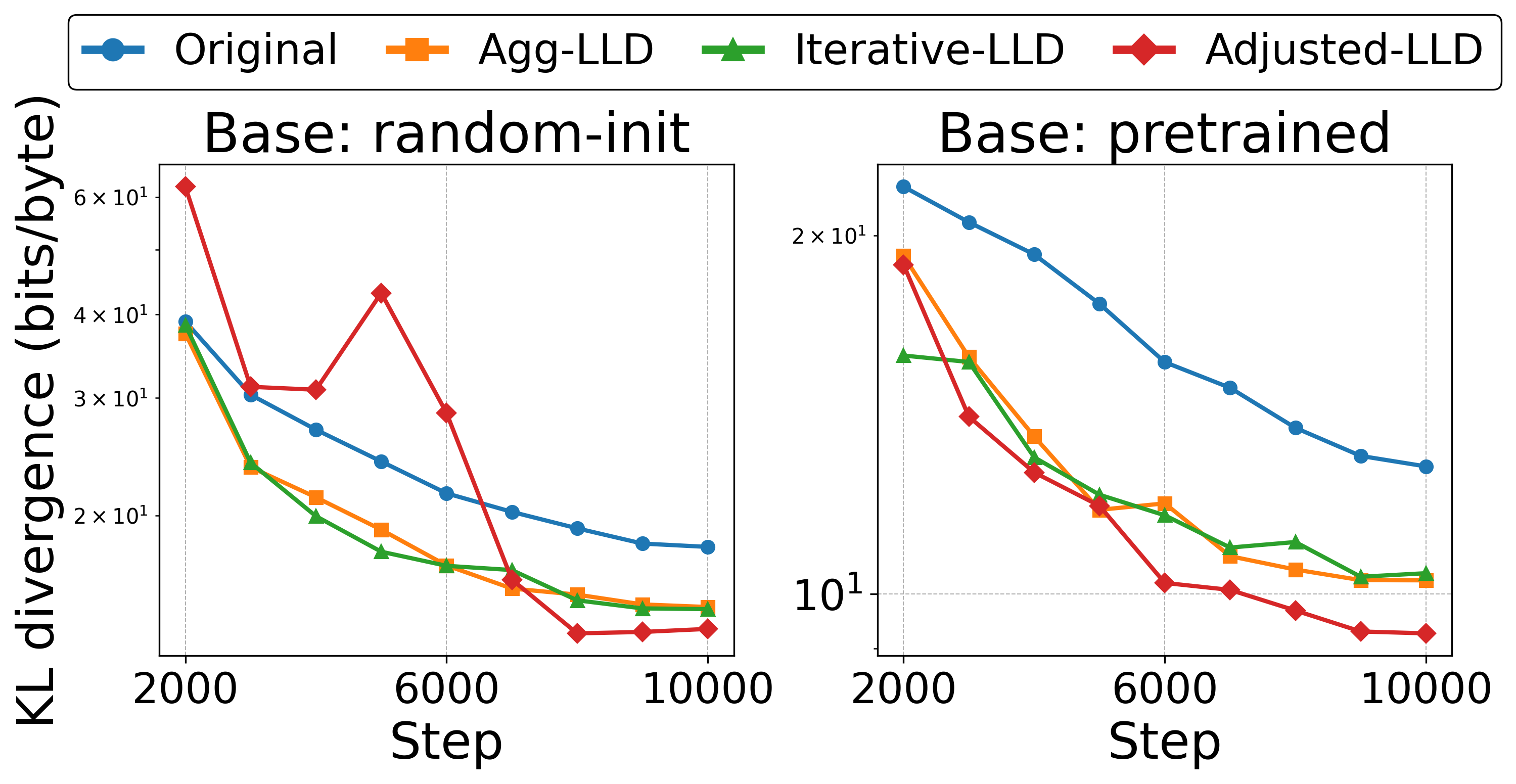}
        \caption{Target model: Gemma-2B.}
        \label{fig:gpt2-gemma}
    \end{subfigure}

    \vspace{0.5em}

    \begin{subfigure}[t]{1.0\linewidth}
        \centering
        \includegraphics[width=1.0\linewidth]{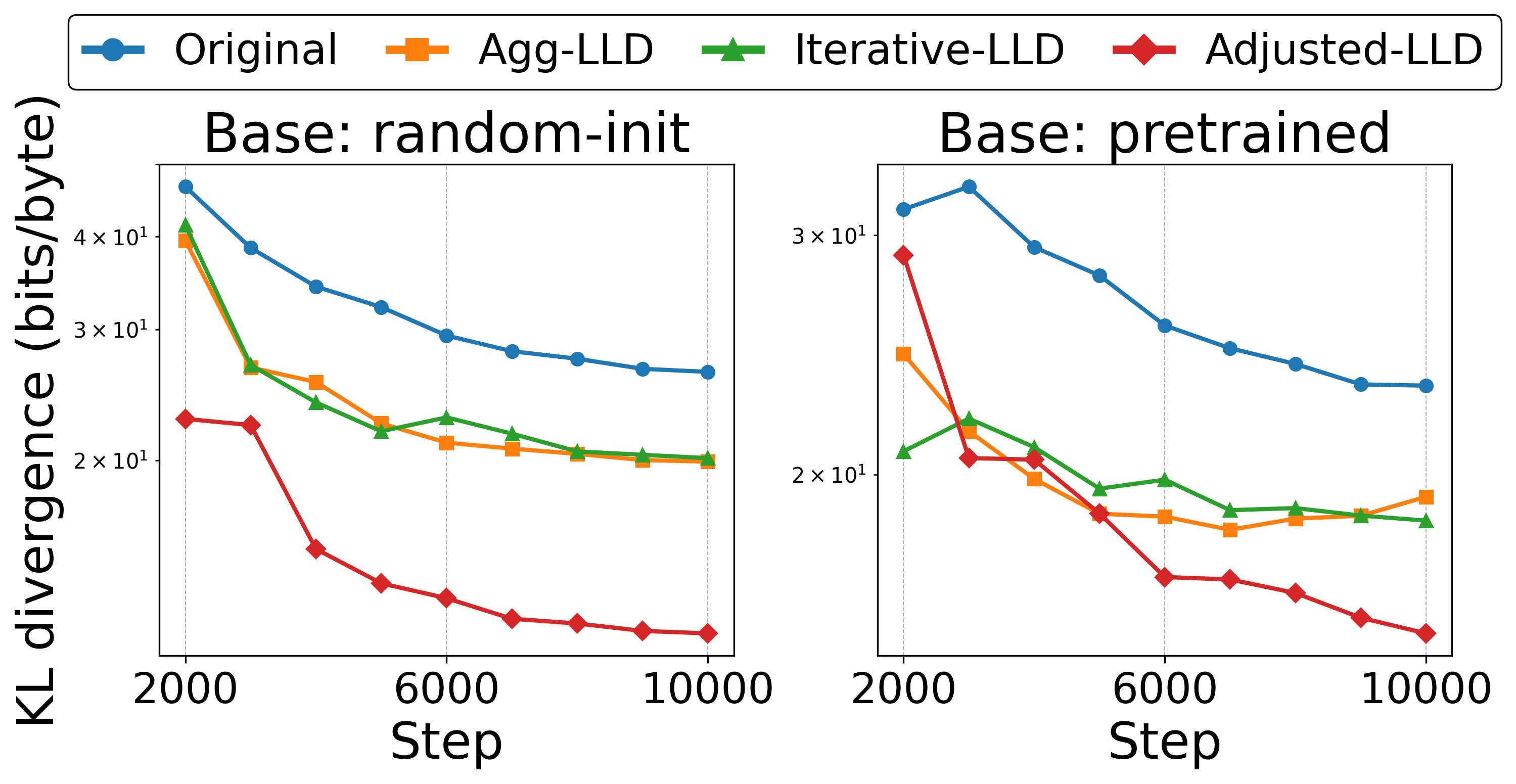}
        \caption{Target model: CodeGemma-2B.}
        \label{fig:gpt2-codegemma}
    \end{subfigure}

    \caption{
    KL divergence between each checkpoint and the target model from step 2k onward for random-init and pretrained NanoGPT trained with different domain weights. Results over all steps are provided in Fig.~\ref{fig:gpt2-kl-all} in Appendix~\ref{app:kl-fullstep}.
    }
    \label{fig:gpt2-kl-main}
\end{figure}

We use Gemma-2B and CodeGemma-2B as target models, and randomly initialized or pretrained NanoGPT as base models. Figs.~\ref{fig:gpt2-gemma} and~\ref{fig:gpt2-codegemma} show KL divergence over training steps, and Table~\ref{tab:ft-kl} reports final-step KL divergence\footnote{For readability, the main plots omit the initial phase of training; the corresponding full curves over all training steps are shown in Appendix~\ref{app:kl-fullstep}.}. Overall, LLD-based domain weights achieve lower KL divergence than Pile-original domain weights in all settings.

Among the three methods, adjusted-LLD serves as the theoretical reference method and indeed achieves the smallest KL divergence in all settings. Although iterative-LLD and aggregated-LLD do not reach adjusted-LLD, both outperform Pile-original weights in all settings. Moreover, the difference between iterative-LLD and aggregated-LLD is small. Therefore, aggregated-LLD has a practical advantage: it provides fixed domain weights set before training while maintaining performance close to iterative-LLD. 

For the pretrained NanoGPT--Gemma-2B setting, Fig.~\ref{fig:traj-original-proposal} visualizes, using t-SNE~\cite{JMLR:v9:vandermaaten08a}, the training checkpoints of pretrained NanoGPT in the \(\bm q\)-coordinates obtained by double-centering text-level log-likelihood vectors on the evaluation corpus, following \citet{oyama-etal-2025-mapping}. By \eqref{eq:KL}, the squared Euclidean distance in this space approximates the KL divergence between models. The figure shows that training with the domain weights estimated by aggregated-LLD moves closer to the target model than training with Pile-original weights.

\begin{table}[t]
\centering
\scriptsize
\begin{tabular}{llcc}
\toprule
\multirow{2}{*}{Target} & \multirow{2}{*}{Method} & \multicolumn{2}{c}{Base} \\
\cmidrule(lr){3-4}
 &  & random-init & pretrained \\
\midrule

\multirow{4}{*}{Gemma-2B}
& Original & 17.9 & 12.8 \\
& Aggregated-LLD & 14.6 & 10.3 \\
& Iterative-LLD & 14.5 & 10.4 \\
& Adjusted-LLD & 13.5 & 9.26 \\
\midrule

\multirow{4}{*}{CodeGemma-2B}
& Original & 26.3 & 23.2 \\
& Aggregated-LLD & 19.9 & 19.3 \\
& Iterative-LLD & 20.1 & 18.5 \\
& Adjusted-LLD & 11.7 & 15.3 \\
\bottomrule
\end{tabular}
\caption{
Final-step KL divergence (bits/byte) to the target model for random-init/pretrained NanoGPT trained with different domain weights.
}
\label{tab:ft-kl}
\end{table}

\paragraph{Robustness to the evaluation corpus.}
In the experiments above, the same evaluation corpus is used both to estimate the domain weights and to compute the approximate KL divergence. To examine whether the observed improvement is specific to this corpus, we reevaluate the final checkpoints on two held-out corpora: another independently sampled set of 10,000 Pile texts and 10,000 texts sampled from SlimPajama~\cite{cerebras2023slimpajama}, both chunked into 1024-byte segments following \citet{oyama-etal-2025-mapping}.
Aggregated-LLD achieves lower KL divergence than Pile-original weighting in all four target/base-model settings, with reductions of 2.2--6.4 bits/byte on the held-out Pile corpus and 1.7--2.5 bits/byte on SlimPajama. These results suggest that the improvement is not specific to the evaluation corpus used in the main experiments.
Full results are provided in Appendix~\ref{app:heldout_eval}.

\paragraph{Empirical step efficiency.}
Using the KL divergence curves in Fig.~\ref{fig:gpt2-kl-main}, we also measure the first step at which aggregated-LLD reaches the KL divergence attained by Pile-original weighting at its 10k-step checkpoint. For Gemma-2B, this occurs at approximately 5.5k and 4.4k steps for the random-init and pretrained base models, respectively; for CodeGemma-2B, the corresponding values are 3.3k and 2.4k steps. These correspond to 45.0\%, 55.7\%, 66.9\%, and 75.7\% fewer steps, respectively. The crossing points are estimated by linear interpolation between checkpoints. Because the trajectories may not yet have converged, we interpret these results as empirical step efficiency rather than as a formal convergence-rate result.

\paragraph{Analysis of \(({\bm J}^\top {\bm J})^{-1}\).}
\begin{figure}[t]
    \centering
    \includegraphics[width=1.0\linewidth]{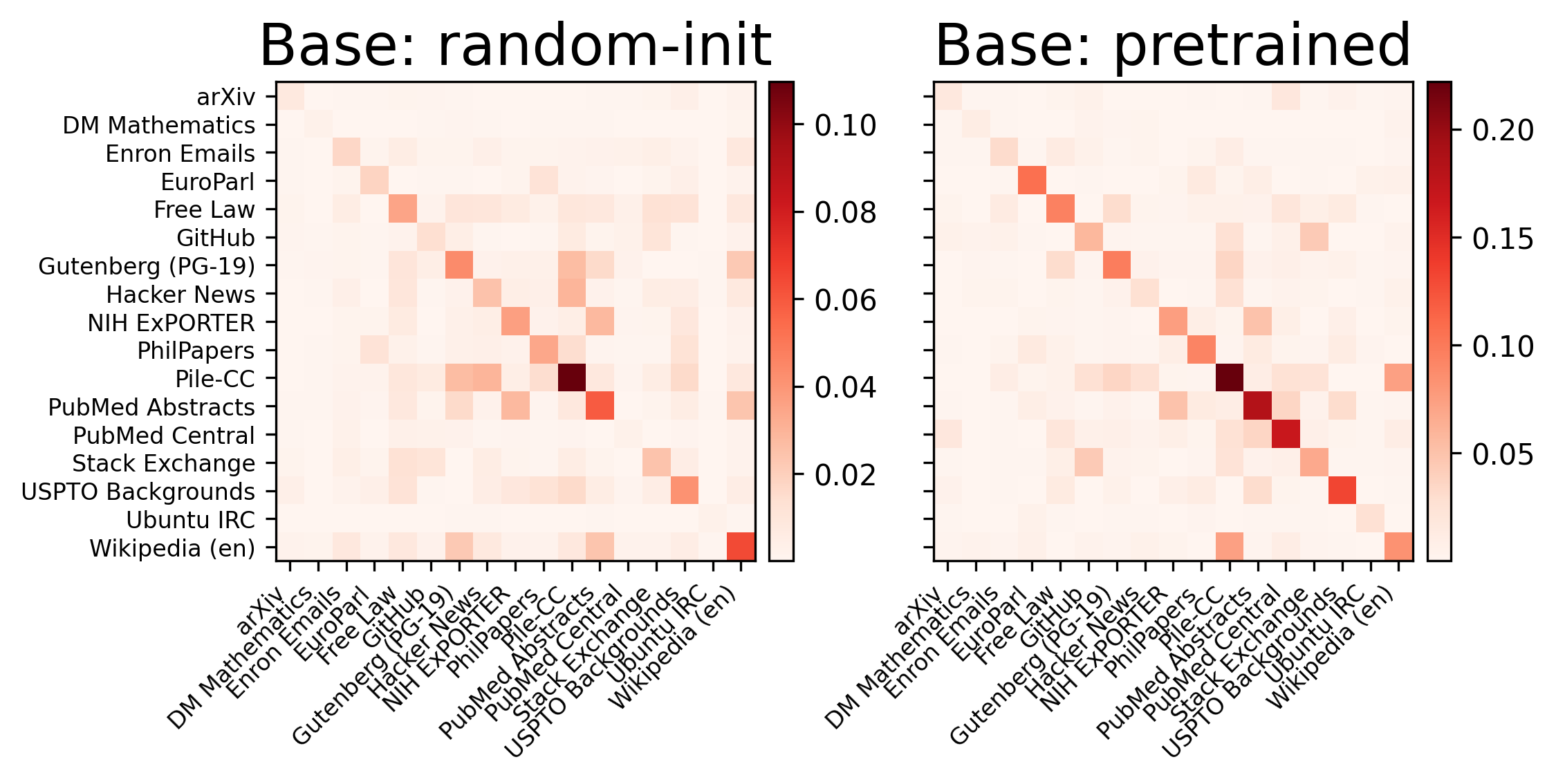}
    \caption{
Heatmaps of the absolute values of the entries of the adjustment matrix \(({\bm J}^\top {\bm J})^{-1}\) for random-init and pretrained NanoGPT, computed on the evaluation corpus $\mathcal{D}$ at step 0. Rows and columns correspond to domains in the Pile.
    }
    \label{fig:gram-matrix}
\end{figure}
Fig.~\ref{fig:gram-matrix} shows the absolute values of the entries of the adjustment matrix \(({\bm J}^\top {\bm J})^{-1}\) for randomly initialized and pretrained NanoGPT, computed on the evaluation corpus \(\mathcal{D}\) at step \(t=0\). In both cases, the diagonal entries are relatively large, indicating that the adjustment is dominated by each domain's own component. This is consistent with the practical effectiveness of LLD without explicit sensitivity adjustment.
For pretrained NanoGPT, however, several off-diagonal entries are also non-negligible, including those between GitHub and Stack Exchange and between Pile-CC and Wikipedia (en). Correcting for such inter-domain correlations may help explain the stronger performance of adjusted-LLD.

\paragraph{Effectiveness of adjusted-LLD.}
As shown in Table~\ref{tab:ft-kl}, adjusted-LLD performs best when its domain weights are updated stage-wise during training. However, when the adjusted-LLD weights are aggregated into a single fixed mixture, the resulting performance is not consistently better than aggregated-LLD. The final-step KL values are 16.6 and 15.9 for Gemma-2B, and 15.2 and 17.7 for CodeGemma-2B, with randomly initialized and pretrained base models, respectively.
Thus, if the goal is to obtain a single fixed training recipe, aggregating the simpler LLD weights without the sensitivity adjustment is a practical choice.

\subsection{Comparison with knowledge distillation}\label{subsec:distill-comparision}

\begin{figure}[t]
    \centering
    \includegraphics[width=1.0\linewidth]{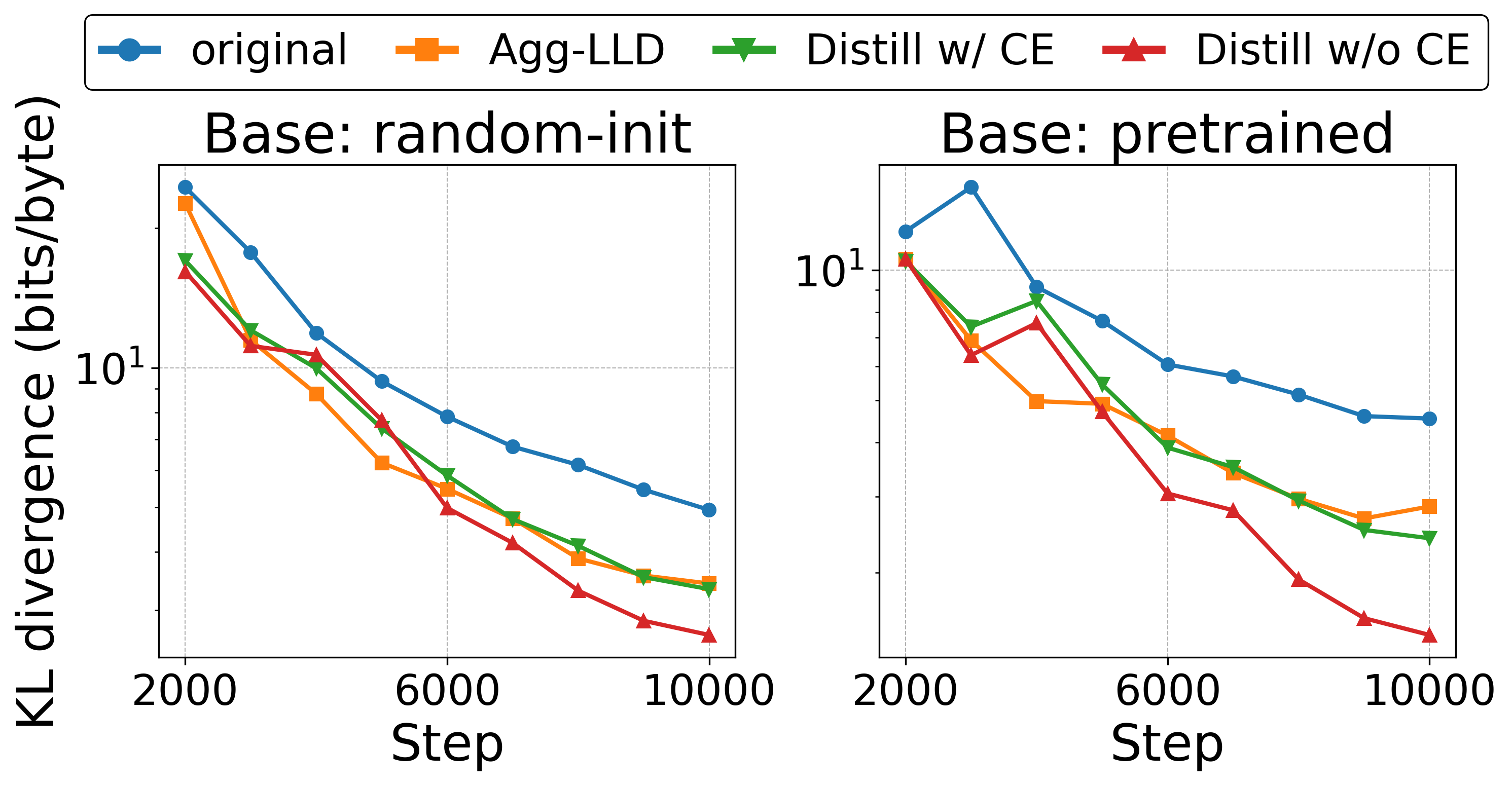}
    \caption{
    KL divergence between each checkpoint and the target model, CodeNanoGPT, from step 2k onward for random-init and pretrained NanoGPT trained with Pile-original domain weights, domain weights estimated by aggregated-LLD, or distillation. Results over all steps are provided in Fig.~\ref{fig:gpt2-distill-all} in Appendix~\ref{app:kl-fullstep}.
    }
    \label{fig:gpt2-distill}
\end{figure}

We compare aggregated-LLD with knowledge distillation. We use CodeNanoGPT as the target model so that the base model and the target model share the same tokenizer. For distillation, we use Pile-original weights and consider two objectives: KL divergence alone, and the sum of KL divergence and the cross-entropy (CE) loss. Although optimizing domain weights for distillation could yield further improvements~\cite{NEURIPS2024_b206d54f}, we adopt this simplest setting to make the comparison with the proposed method clearer.

The results are shown in Fig.~\ref{fig:gpt2-distill}. When randomly initialized NanoGPT is used as the base model, the KL divergence from the target model at the final training step is 4.93, 3.42, 3.33, and 2.65 bits/byte for Pile-original, aggregated-LLD, distill with CE, and distill without CE, respectively. When pretrained NanoGPT is used as the base model, the corresponding values are 4.54, 2.84, 2.40, and 1.43 bits/byte. Fig.~\ref{fig:traj-original-proposal} also visualizes the corresponding training trajectories in the log-likelihood space.
As expected, distillation without CE yields the lowest KL divergence, since it directly optimizes the KL-based objective. However, aggregated-LLD remains close to distillation with CE, even though it uses only pre-estimated domain weights and does not access teacher outputs during training. This indicates that the proposed method can recover much of the alignment achieved by distillation with CE and be a practical distillation-free alternative when teacher outputs are unavailable.

\subsection{Similarity in task performance}\label{subsec:eval}

\begin{figure}[t]
\centering
\begin{subfigure}{0.98\linewidth}
\centering
\scriptsize
\setlength{\tabcolsep}{3.0pt}
\renewcommand{\arraystretch}{1.08}
\begin{tabular}{lccccc}
\toprule
Model & ARC-Easy & HellaSwag & TruthfulQA & PIQA & LAMBADA \\
\midrule
Target & 0.426 & 0.283 & 0.259 & 0.618 & 0.284 \\
random-init & 0.358 & 0.267 & 0.272 & 0.559 & 0.113 \\
pretrained & 0.425 & 0.283 & 0.247 & 0.613 & 0.292 \\
\bottomrule
\end{tabular}
\caption{Final-step accuracy of the target model and the two trained base models on five downstream tasks.}
\label{fig:downstream_acc_table}
\end{subfigure}
\vspace{0.5em}

\begin{subfigure}{0.98\linewidth}
\centering
\includegraphics[width=1.0\linewidth]{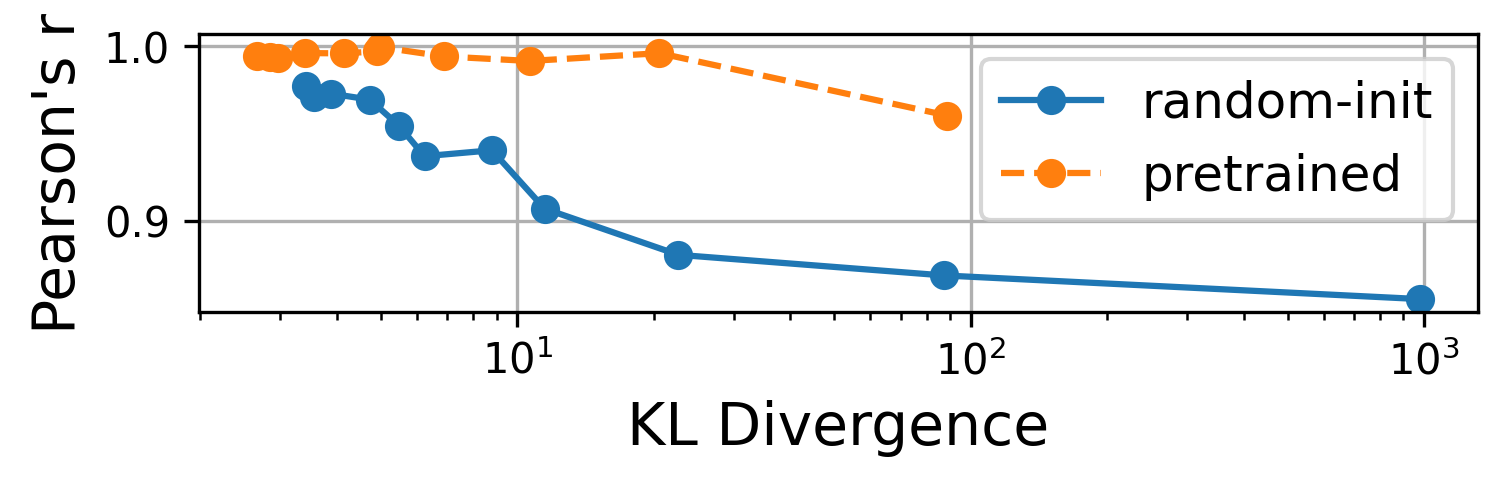}
\caption{
Relationship across training checkpoints between the KL divergence to the target model and the Pearson correlation between the five-task performance profile of each checkpoint and that of the target model.
}
\label{fig:downstream_corr}
\end{subfigure}
\caption{
 Downstream task evaluation with CodeNanoGPT as the target model and randomly initialized or FineWeb-pretrained NanoGPT as the base model, trained using aggregated-LLD.
}
\label{fig:evaluation}
\end{figure}

We examine whether bringing a model closer to the target model in log-likelihood space using the proposed method also makes its downstream performance profile closer to that of the target model.

We use randomly initialized and FineWeb-pretrained NanoGPT as the base models and CodeNanoGPT as the target model. For evaluation, we use AI2 Reasoning Challenge (ARC)~\cite{clark2018thinksolvedquestionanswering}, HellaSwag~\cite{zellers-etal-2019-hellaswag}, TruthfulQA~\cite{lin-etal-2022-truthfulqa}, PIQA~\cite{bisk2019piqareasoningphysicalcommonsense}, and LAMBADA~\cite{paperno-etal-2016-lambada}, inspired by evaluations of small language models~\cite{NEURIPS2024_370df50c}.
All evaluations are conducted using \texttt{lm-evaluation-harness}~\cite{eval-harness}.

As shown in Fig.~\ref{fig:evaluation}, the final checkpoint trained with aggregated-LLD exhibits task performance that generally corresponds to that of the target model across downstream tasks. Moreover, as the KL divergence to the target model decreases, the Pearson correlation between the performance vectors also tends to increase\footnote{Note that, in the multiple-choice tasks other than LAMBADA, correlations can arise even at initialization because of the effect of chance rates.}.

We also conduct a likelihood-based downstream analysis using BPB on correct completions, following the motivation that accuracy can be insensitive for small models~\cite{heineman2025signalnoiseframeworkreducing}. The results show a similar tendency: the proposed method makes the downstream likelihood profiles closer to those of the target model, especially in the random-init setting. Details are provided in Appendix~\ref{app:downstream_bpb}.

\section{Conclusion}\label{sec:conclusion}

We proposed a method for distributionally aligning a base model with a target model during pretraining or continued pretraining by controlling domain weights in the training data. Using a geometric formulation in log-likelihood space, we derived a theoretical criterion for domain weights that steer the update direction toward the target model and approximated it with fixed domain weights. The resulting mixture can be fixed before the second-pass training, allowing alignment to be carried out as an ordinary training run without further access to the target model. Experiments with NanoGPT showed that the proposed method yields smaller KL divergence than Pile-original domain weights and brings the base model closer to the target model. For target models that share the same tokenizer, the proposed method provides a useful alternative to knowledge distillation. As KL divergence decreases, downstream task performance also tends to become closer to that of the target model.

\section*{Limitations}
\begin{itemize}
    \item Our experiments are conducted primarily using NanoGPT with 124M parameters. Whether the same behavior also holds for substantially larger models, such as those with several billion parameters or more, has not been examined due to computational resource constraints and remains future work.

    \item We primarily evaluate our method on specific pretrained models. Its generality to a broader range of architectures, as well as to models that have undergone instruction tuning or alignment through reinforcement learning, has not yet been sufficiently examined.
    
    \item This study assumes the Pile Uncopyrighted and thus relies on a large-scale corpus with explicitly defined domains.
    The proposed framework is not specific to the Pile, and could be applied to other corpora with predefined partitions, such as RedPajama~\cite{NEURIPS2024_d3449733}, or WebOrganizer~\cite{wettig2025organize}. 
    It also remains unverified whether the same optimization of domain weights is effective for other corpora or for datasets with ambiguous domain boundaries. In particular, the results may vary substantially depending on the granularity of the domain partitioning. Extending the method to finer granularities, such as sentence-level units, to pseudo-domains obtained by clustering in embedding space~\cite{diao2025nemotronclimbclusteringbasediterativedata}, or to settings in which a text belongs to multiple domains remains important future work.

    \item As described in Section~\ref{subsec:experimental-setup} and Appendix~\ref{app:experimental-setup-detail}, the evaluation text set $\mathcal{D}$ is the same as that used by \citet{oyama-etal-2025-mapping} and therefore consists of texts with variable token lengths. Since actual training is performed with a fixed token length, we normalize the log-likelihood by token length for each domain. Ideally, such normalization would be unnecessary if a fixed-token-length evaluation set were used from the outset. At the same time, allowing variable-length texts is itself natural when users may evaluate models on arbitrary text sets. The effect of this particular choice of $\mathcal{D}$ on the results has not been examined.

    \item For simplicity, the theory assumes SGD as the training algorithm, whereas the experiments use AdamW, which is widely adopted in practice. Although we empirically confirm the alignment effect in this study, the exact correspondence between the internal dynamics of AdamW and the geometric interpretation in log-likelihood space remains unclear.

    \item Adjusted-LLD uses the adjustment matrix $(\bm{J}^\top \bm{J})^{-1}$, but computing it requires $O(K^2P)$ time, making both the computational and memory costs substantial for large language models. For this reason, we treat the criterion based on $(\bm{J}^\top \bm{J})^{-1}$ as a theoretical reference method and primarily use its approximations, iterative-LLD and aggregated-LLD, in our experiments. Since what is actually required is the Gram matrix $\bm{J}^\top \bm{J}$ rather than the Jacobian itself, approximate computation via partial parameter sampling or stochastic estimation remains an important direction for future work.

    \item In \eqref{eq:raw-domain-ratio}, it is also possible to vary $\tau$ across training steps. In particular, since the likelihood is expected to fluctuate substantially in the early stage of training the base model, one possible strategy is to use a higher temperature at the beginning and gradually lower it as training proceeds. It is also conceivable to automatically scale the temperature using the variance of the domain-level log-likelihood differences across domains. In addition, in \eqref{eq:domain-ratio-aggregate}, when aggregating the mixture proportions estimated at each step, one could consider taking a weighted average rather than an unweighted one. Improving the method along these lines remains future work.

    \item Our method estimates domain weights for distributional alignment to a target model, not for directly optimizing domain-wise performance. Thus, the estimated weights should not be interpreted as predicting per-domain performance improvements. A natural extension is to optimize mixture weights with respect to an explicit objective over domain-wise validation losses or downstream task metrics, possibly using proxy-based mixture optimization frameworks such as RegMix~\cite{liu2025regmix}.
\end{itemize}

\section*{Ethical Considerations}

We note a potential misuse risk: because the first pass references a target model but the second pass can be executed without further access to it, the role of the target model may become obscured. This method should not be used to conceal target-model-dependent training or to circumvent model licenses or usage restrictions. When applying the method, practitioners should disclose whether domain weights were estimated using a target model and ensure that such use complies with the target model's license and terms of use.

\section*{Acknowledgments}

This work was partially supported by JSPS KAKENHI JP22H05106, JP23K28045, JP26H02495 (to HS), JP25K24366 (to HY), JP26KJ1485 (to MO), JST CREST JPMJCR21N3 (to HS), JST Moonshot JPMJMS263I (to HS), and JST BOOST JPMJBS2407 (to MO).

\bibliography{custom, model_list}

\appendix
\section{Mathematical Details}\label{app:theory}

\subsection{Proof of (\ref{eq:sgd-logp-diff})}\label{app:proof-sgd-logp-diff}

We show how the coordinates in log-likelihood space change under parameter updates by SGD:
\[
    \mathbb{E}_{x\sim r_{\bm \pi}}[{\bm\ell}(\bm\theta+\Delta\bm\theta) - {\bm\ell}(\bm\theta)] 
    \approx \eta{\bm J}(\bm\theta)^\top {\bm J}(\bm\theta)\bm\pi .
\]

First, the expected parameter update direction is approximated as follows:
\begin{align*}
    &\hspace{-1.5em}\mathbb{E}_{x\sim r_{\bm \pi}}[\Delta\bm\theta]\\
    &= \mathbb{E}_{x\sim r_{\bm \pi}}\left[\eta\nabla_{\bm\theta}\log p_{\bm \theta}(x)\right] \\
    &= \sum_{k=1}^K \pi_k \mathbb{E}_{x\sim r_k}\left[\eta\nabla_{\bm\theta}\log p_{\bm \theta}(x)\right] \\
    &\approx \eta\sum_{k=1}^K \pi_k\nabla_{\bm\theta}\left(\frac{1}{|\mathcal{D}_k|}\sum_{x\in\mathcal D_k}\log p_{\bm \theta}(x)\right) \\
    &= \eta {\bm J}(\bm\theta) \bm\pi .
\end{align*}
In the last line, we use the fact that the $k$-th column of the Jacobian ${\bm J}(\bm\theta)$ is
\[
 \nabla_{\bm\theta} \left(\frac{1}{|\mathcal{D}_k|}\sum_{x\in\mathcal D_k}\log p_{\bm \theta}(x)\right).
\]
The approximation above replaces the expectation over $x\sim r_k$ with the sample mean over the evaluation text set $\mathcal{D}_k$. This is justified under the assumption that $\mathcal{D}_k$ is sampled from $r_k$.

Next, we approximate the change in the domain-level log-likelihood vector induced by a small parameter update using the first-order Taylor expansion with respect to $\bm\theta$:
\begin{align*}
    {\bm\ell} (\bm\theta + \Delta \bm\theta) \approx {\bm\ell}(\bm\theta) + {\bm J}(\bm\theta)^\top \Delta\bm\theta .
\end{align*}
Therefore, we obtain
\begin{align*}
    \mathbb{E}_{x\sim r_{\bm \pi}}[{\bm\ell}(\bm\theta+\Delta\bm\theta) - {\bm\ell}(\bm\theta)]
    &\approx \mathbb{E}_{x\sim r_{\bm\pi}}[{\bm J}(\bm\theta)^\top \Delta\bm\theta] \\
    &\approx \eta{\bm J}(\bm\theta)^\top {\bm J}(\bm\theta)\bm\pi .
\end{align*}

\subsection{Proof of (\ref{eq:opt-domain-ratio})}\label{app:proof-opt-domain-ratio}

We show that the solution to the optimization problem
\begin{align*}
    \max_{\bm\pi\in\Delta^{K-1}}&~\bm \pi^\top \tilde{\bm\pi} - \tau \KL(\bm\pi, \bm\pi')
\end{align*}
is given by
\begin{equation*}
    \pi_k = \frac{\pi'_k\exp(\tilde\pi_k / \tau)}{\sum_{k'=1}^K \pi_{k'}' \exp(\tilde\pi_{k'}/\tau) } .
\end{equation*}

Let $\lambda$ be the Lagrange multiplier for the equality constraint $\sum_k \pi_k = 1$, and define
\begin{align*}
\mathcal{L}(\bm\pi,\lambda)
& =
\sum_{k=1}^K \pi_k \tilde\pi_k
-\tau \sum_{k=1}^K \pi_k\log\frac{\pi_k}{\pi_k'}\\
&\quad+\lambda\left(\sum_{k=1}^K \pi_k-1\right).
\end{align*}
Taking the partial derivative with respect to $\pi_k$ gives
\[
\frac{\partial \mathcal{L}}{\partial \pi_k}
=
\tilde\pi_k
-\tau\left(\log\frac{\pi_k}{\pi_k'}+1\right)
+\lambda.
\]
Therefore, at the optimum, we have
\[
\tilde\pi_k-\tau\left(\log\frac{\pi_k}{\pi_k'}+1\right)+\lambda=0.
\]
Rearranging this expression yields
\[
\log\frac{\pi_k}{\pi_k'}
=
\frac{\tilde\pi_k+\lambda-\tau}{\tau}.
\]
Hence,
\begin{align*}
\pi_k &= \pi_k' \exp\!\left(\frac{\tilde\pi_k+\lambda-\tau}{\tau}\right)\\
&= \pi_k' \exp\!\left(\frac{\tilde\pi_k}{\tau}\right)\exp\!\left(\frac{\lambda-\tau}{\tau}\right)\\
&\propto \pi_k' \exp({\tilde\pi_k}/{\tau}) > 0.
\end{align*}
Normalizing over $k$, we obtain
\[
\pi_k
=
\frac{\pi_k' \exp(\tilde\pi_k/\tau)}
{\sum_{k'=1}^K \pi_{k'}' \exp(\tilde\pi_{k'}/\tau)}.
\]

\subsection{Relation to mirror descent}\label{app:mirror-descent}

Let us define the distance-generating function as
\[
h(\bm\pi) = \sum_{k=1}^K \pi_k \log\pi_k,
\]
so the primal variable $\bm\pi$ is mapped to the dual variable $\bm z = \nabla h(\bm \pi)$, whose components are
\[
(\nabla h(\bm\pi))_k = \log \pi_k + 1.
\]
Likewise, the reference domain weight vector $\bm\pi'$ is mapped to $\bm z' = \nabla h(\bm \pi')$, whose components are
\[
z'_k = \log \pi_k' + 1 .
\]
Since the gradient of the linear term in the primal objective \eqref{eq:reg-problem} is $\tilde{\bm\pi}$, one mirror-descent step in the dual space with step size $1/\tau$ is
\[
\bm z = \bm z' + \frac{1}{\tau}\tilde{\bm\pi} .
\]
Mapping this back to the primal space, we have $\bm\pi = \nabla h^*(\bm z)$, and
\begin{align*}
{\pi}_k &\propto \exp(z_k-1)\\
&= \exp\left(\log\pi_k' + \frac{\tilde\pi_k}{\tau} \right)\\
&= \pi_k'\exp({\tilde\pi_k}/{\tau}) .
\end{align*}
Therefore, the update in the primal space is
\[
    \pi_k = \frac{\pi'_k\exp(\tilde\pi_k / \tau)}{\sum_{k'=1}^K \pi_{k'}' \exp(\tilde\pi_{k'}/\tau) } .
\]
Thus, our update can be interpreted as taking one mirror-descent step from $\bm\pi'$.

\subsection{Proof of (\ref{eq:domain-ratio-aggregate-optimization})}\label{app:proof-domain-ratio-aggregate}

We show that the solution to the optimization problem
\[
   \min_{\bm\pi \in \Delta^{K-1}}\sum_{t\in\mathcal T} \KL(\bm\pi,\bm\pi^{(t)})
\]
is given by
\[
    \pi^*_k \propto \left(\prod_{t\in\mathcal T} \pi^{(t)}_k\right)^{1/|\mathcal T|} .
\]

Let $\lambda$ be the Lagrange multiplier for the constraint $\sum_k \pi_k = 1$, and define
\begin{align*}
 \mathcal L(\bm\pi,\lambda)
 &=
 \sum_{t\in\mathcal T} \KL(\bm\pi,\bm\pi^{(t)}) +
 \lambda\left(\sum_{k=1}^K\pi_k-1\right)\\
 &=
\sum_{k=1}^K \left(|\mathcal T|\pi_k\log\pi_k-\pi_k\sum_{t\in\mathcal T}\log\pi_k^{(t)}\right) \\
&\quad+\lambda\left(\sum_{k=1}^K\pi_k-1\right).
\end{align*}
Taking the partial derivative with respect to $\pi_k$, we obtain
\[
\frac{\partial \mathcal L}{\partial \pi_k}
=
|\mathcal T|(\log\pi_k+1)-\sum_{t\in\mathcal T}\log\pi_k^{(t)}+\lambda.
\]
Therefore, at the optimum, we have
\[
|\mathcal T|(\log\pi_k+1)-\sum_{t\in\mathcal T}\log\pi_k^{(t)}+\lambda=0.
\]
Rearranging this, we obtain
\[
\log\pi_k
=
\frac{1}{|\mathcal T|}\sum_{t\in\mathcal T}\log\pi_k^{(t)}
-\frac{\lambda+|\mathcal T|}{|\mathcal T|}.
\]
That is,
\[
\pi_k^* \propto \exp\left( \frac{1}{|\mathcal T|}\sum_{t\in\mathcal T}\log\pi_k^{(t)} \right).
\]
Normalizing over $k$, we obtain
\[
\pi_k^*
=
\frac{\left(\prod_{t\in\mathcal T}\pi_k^{(t)}\right)^{1/|\mathcal T|}}
{\sum_{j=1}^K \left(\prod_{t\in\mathcal T}\pi_j^{(t)}\right)^{1/|\mathcal T|}}.
\]

\section{Details of Experimental Setup}\label{app:experimental-setup-detail}

\subsection{Datasets}
\begin{table}[t]
\centering
\small
\begin{tabular}{lrr}
\toprule
Domain & \# Texts & Mean tokens\\
\midrule
arXiv & 1172 & 365 \\
DM Mathematics & 151 & 481 \\
Enron Emails & 22 & 242 \\
EuroParl & 83 & 416 \\
Free Law & 837 & 262 \\
GitHub & 925 & 429 \\
Gutenberg (PG-19) & 251 & 270 \\
Hacker News & 67 & 248 \\
NIH ExPORTER & 41 & 179 \\
PhilPapers & 51 & 256 \\
Pile-CC & 2353 & 220 \\
PubMed Abstracts & 464 & 179 \\
PubMed Central & 1763 & 317 \\
Stack Exchange & 712 & 311 \\
USPTO Backgrounds & 487 & 199 \\
Ubuntu IRC & 54 & 406 \\
Wikipedia (en) & 567 & 217 \\
\bottomrule
\end{tabular}
\caption{
Number of texts and mean token length for each domain in the evaluation corpus $\mathcal{D}$.
}
\label{tab:the-pile-domain-numtexts}
\end{table}

\begin{table}[t]
\centering
\scriptsize
\begin{tabular}{lrrr}
\toprule
Domain & Original & Code & PubMed \\
\midrule
arXiv & 0.119 & 0.0356 & 0.0107 \\
DM Mathematics & 0.0165 & 0.00493 & 0.00148 \\
Enron Emails & 0.00187 & 0.000557 & 0.000167 \\
EuroParl & 0.00973 & 0.0029 & 0.000869 \\
Free Law & 0.0816 & 0.0243 & 0.00729 \\
GitHub & 0.101 & \textbf{0.405} & 0.00904 \\
Gutenberg (PG-19) & 0.0289 & 0.00863 & 0.00258 \\
Hacker News & 0.00826 & \textbf{0.0331} & 0.000738 \\
NIH ExPORTER & 0.004 & 0.00119 & 0.000357 \\
PhilPapers & 0.00507 & 0.00151 & 0.000452 \\
Pile-CC & 0.241 & 0.072 & 0.0216 \\
PubMed Abstracts & 0.0409 & 0.0122 &\textbf{ 0.164} \\
PubMed Central & 0.192 & 0.0573 & \textbf{0.768} \\
Stack Exchange & 0.0684 & \textbf{0.274} & 0.00611 \\
USPTO Backgrounds & 0.0487 & 0.0145 & 0.00435 \\
Ubuntu IRC & 0.0117 &\textbf{ 0.0469} & 0.00105 \\
Wikipedia (en) & 0.0204 & 0.00608 & 0.00182 \\
\bottomrule
\end{tabular}

\caption{
Domain weights in the Pile Uncopyrighted. In the Code setting, the weights for GitHub, Stack Exchange, Ubuntu IRC, and Hacker News are quadrupled; in the PubMed setting, the weights for PubMed Abstracts and PubMed Central are quadrupled. The corresponding values are shown in bold.
}
\label{tab:the-pile-domain-ratio}
\end{table}

For the domain names in the Pile, we standardized several labels for readability and consistency. Specifically, ArXiv was changed to arXiv, FreeLaw to Free Law, Github to GitHub, HackerNews to Hacker News, NIH ExPorter to NIH ExPORTER, and StackExchange to Stack Exchange. These changes affect only the presentation of the labels and do not alter the underlying categorization.

We use the evaluation corpus $\mathcal{D}$ from \citet{oyama-etal-2025-mapping} without modification. This corpus consists of texts sampled from the Pile, and Table~\ref{tab:the-pile-domain-numtexts} reports the number of texts and the mean token length for each domain.

In addition, the model used in Fig.~\ref{fig:domain-effect} in Section~\ref{sec:related_work} and CodeNanoGPT used in Section~\ref{sec:experiments} were both trained from NanoGPT with manually adjusted domain weights on the Pile, in which the weights of particular domains were increased. The corresponding domain weights are shown in Table~\ref{tab:the-pile-domain-ratio}. Note that ``Original'' is obtained by taking the 17 uncopyrighted domain weights from the 22 domain weights reported in the original Pile paper~\cite{gao2020pile800gbdatasetdiverse} and renormalizing them.

\subsection{Training environment and hyperparameters}
The training environment and hyperparameters for the base model are as follows. We use the GPT-2 tokenizer. The NanoGPT (124M) used as the base model has a vocabulary size of 50,304, 12 layers, 12 attention heads, an embedding dimension of 768, and a maximum sequence length of 1,024. The total number of tokens per step is 524,288, and the micro-batch size is 64. We use AdamW for optimization, with weight decay 0.1, betas $(0.9, 0.95)$, and epsilon $1\times10^{-8}$. For the learning rate schedule, we adopt cosine decay, with the first 10\% of the total training steps used for warmup, and decay the learning rate from $6\times10^{-4}$ to $6\times10^{-5}$. The total number of training steps is 10k for base model training and knowledge distillation, 100k for CodeNanoGPT, and approximately 190k for pretraining on FineWeb 100BT. Training is conducted using four NVIDIA RTX A6000 Ada GPUs. Under these settings, training for 10k steps takes approximately five hours.

\section{KL Divergence Curves over All Training Steps}\label{app:kl-fullstep}
\begin{figure}[t]
    \centering
    \begin{subfigure}[t]{1.0\linewidth}
        \centering
        \includegraphics[width=1.0\linewidth]{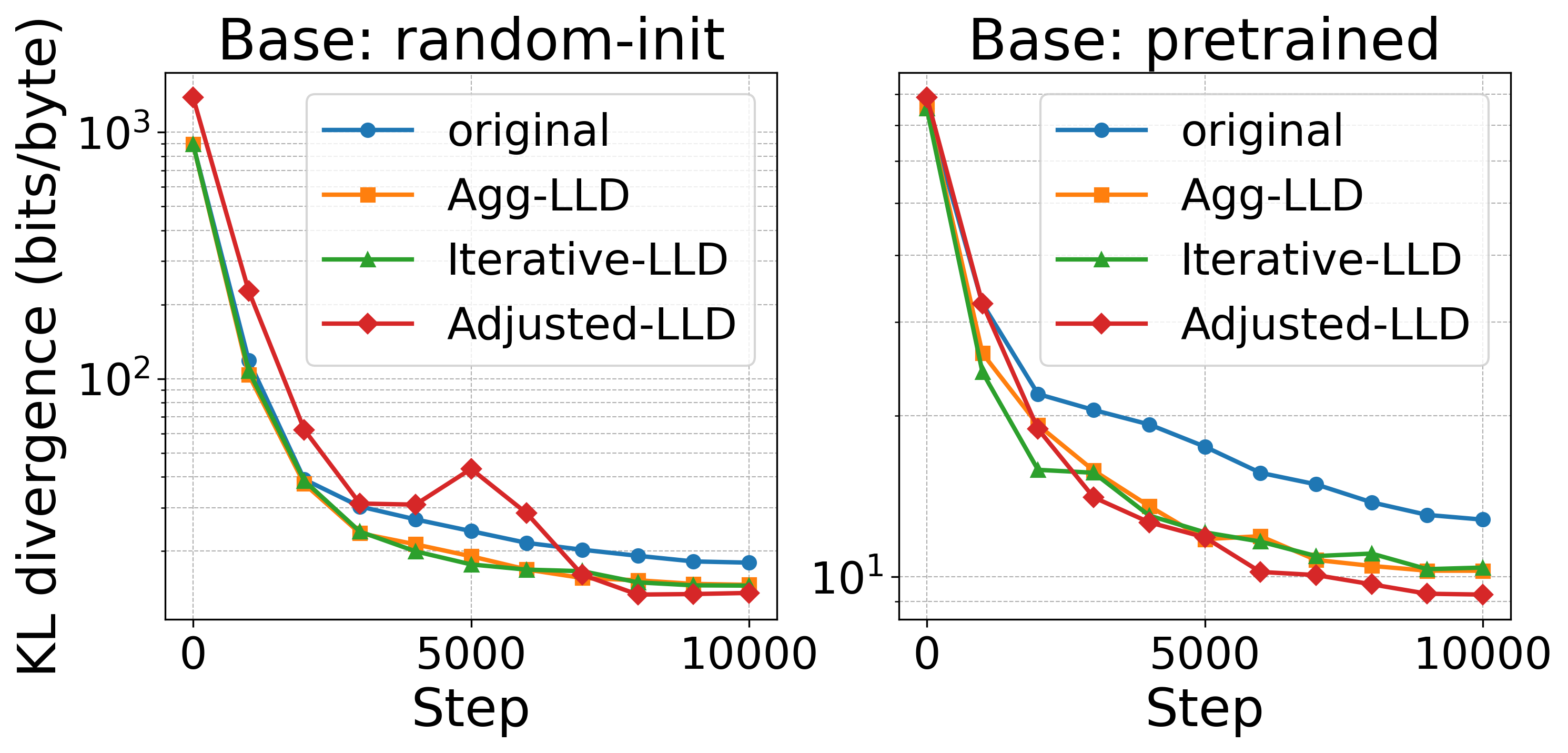}
        \caption{Target model: Gemma-2B.}
        \label{fig:gpt2-gemma-all}
    \end{subfigure}

    \vspace{0.5em}

    \begin{subfigure}[t]{1.0\linewidth}
        \centering
        \includegraphics[width=1.0\linewidth]{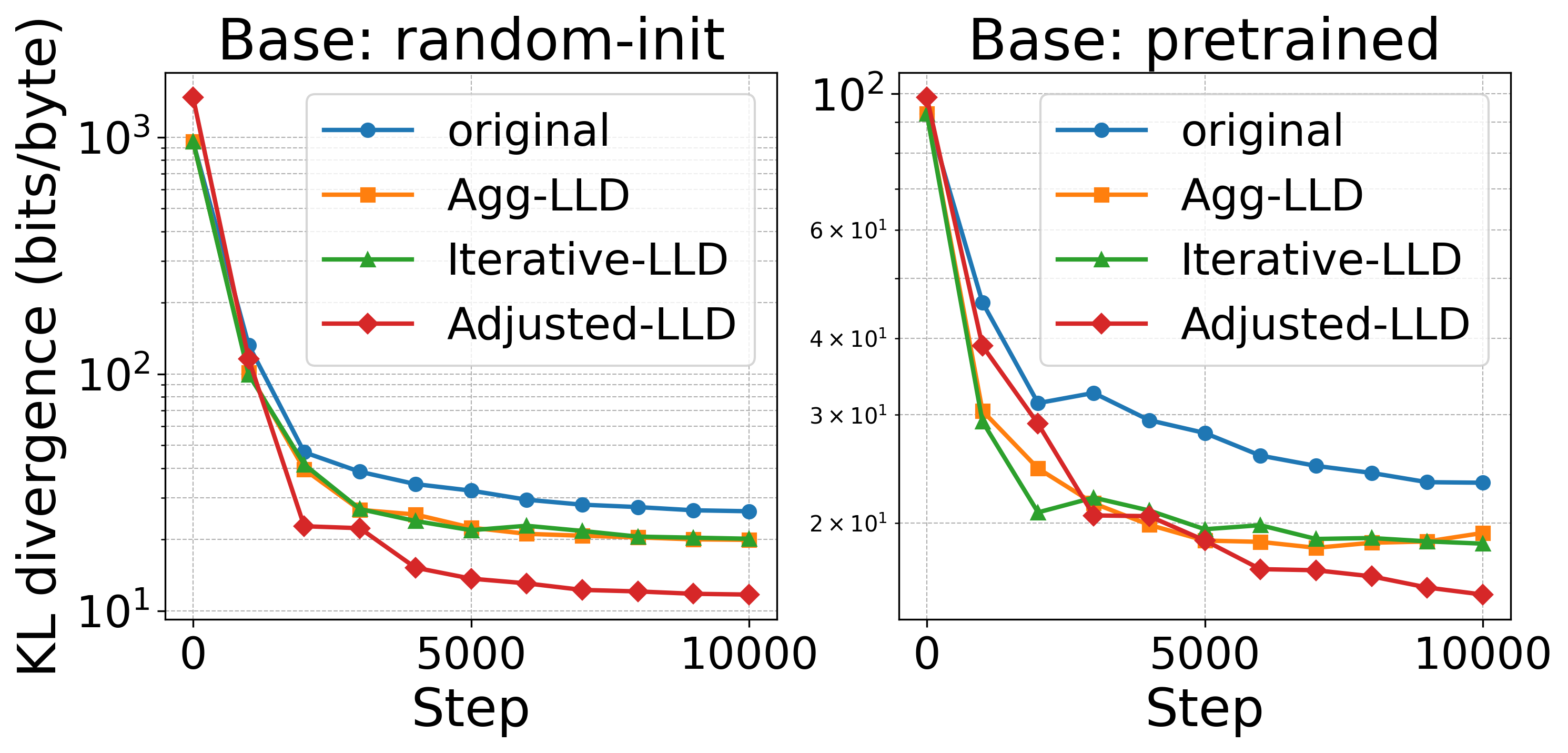}
        \caption{Target model: CodeGemma-2B.}
        \label{fig:gpt2-codegemma-original}
    \end{subfigure}

    \caption{
    Full KL-divergence trajectories corresponding to Fig.~\ref{fig:gpt2-kl-main}, shown over all training steps. The target models are Gemma-2B and CodeGemma-2B, and the base models are random-init or pretrained NanoGPT.
    }
    \label{fig:gpt2-kl-all}
\end{figure}

\begin{figure}[t]
    \centering
    \includegraphics[width=1.0\linewidth]{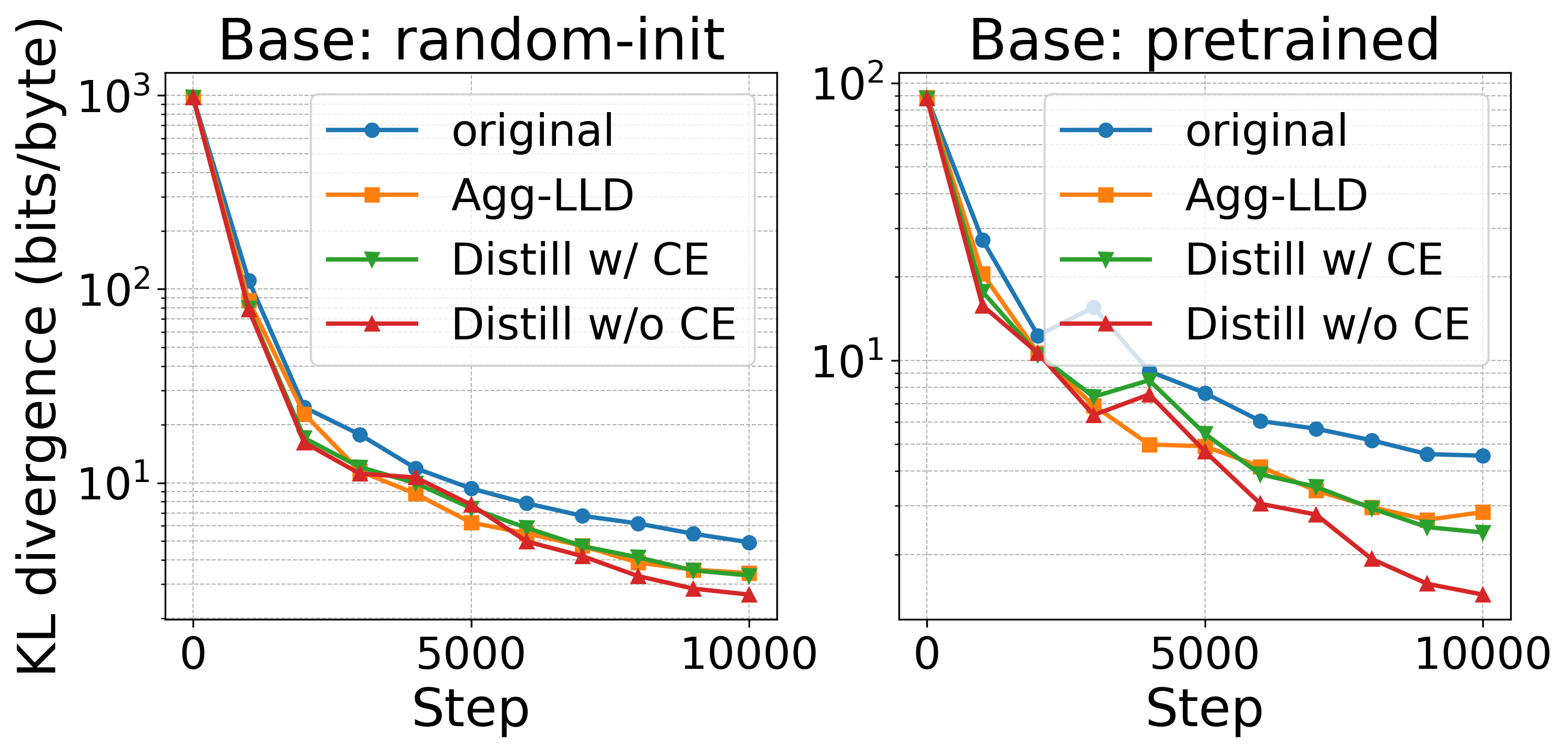}
    \caption{
    Full KL-divergence trajectories corresponding to Fig.~\ref{fig:gpt2-distill}, shown over all training steps. The target model is CodeNanoGPT, and the curves compare Pile-original weighting, aggregated-LLD, and distillation-based training for random-init and pretrained NanoGPT.
    }
    \label{fig:gpt2-distill-all}
\end{figure}

This appendix provides the full KL-divergence curves corresponding to the truncated plots in the main text. Whereas Figs.~\ref{fig:gpt2-kl-main} and \ref{fig:gpt2-distill} focus on checkpoints from step 2k onward for readability, Figs.~\ref{fig:gpt2-kl-all} and \ref{fig:gpt2-distill-all} show the same experiments over all training steps, including the initial phase.

\section{Robustness to Evaluation Corpus}
\label{app:heldout_eval}

In the experiment in Section~\ref{subsec:ft-kl-target}, the same evaluation corpus is used for estimating the domain weights and computing the approximate KL divergence between the trained base model and the target model. 
To examine whether the observed improvements depend on this particular evaluation corpus, we consider two held-out evaluation corpora.

The first is another set of 10,000 texts sampled from the Pile, which follows the same source distribution as the original evaluation corpus but contains different texts. As shown in Table~\ref{tab:ft-kl-other-pile-sample}, aggregated-LLD achieves lower approximate KL divergence than the Pile-original baseline in all target/base-model settings.

Second, we evaluate the final checkpoints on 10,000 texts sampled from SlimPajama\footnote{\url{https://huggingface.co/datasets/DKYoon/SlimPajama-6B}}. Unlike the first evaluation, this corpus is not sampled from the Pile, and thus provides a stronger test of robustness to the choice of evaluation corpus. As shown in Table~\ref{tab:ft-kl-other-redpajama}, aggregated-LLD again achieves lower approximate KL divergence than the Pile-original baseline in all settings.

\begin{table}[t]
\centering
\scriptsize
\begin{tabular}{llcc}
\toprule
\multirow{2}{*}{Target} & \multirow{2}{*}{Method} & \multicolumn{2}{c}{Base} \\
\cmidrule(lr){3-4}
 &  & random-init & pretrained \\
\midrule

\multirow{2}{*}{Gemma-2B}
& Original & 18.5 & 13.0 \\
& Aggregated-LLD & 15.4 & 10.8 \\
\midrule

\multirow{2}{*}{CodeGemma-2B}
& Original & 26.5 & 23.4 \\
& Aggregated-LLD & 20.1 & 19.0 \\
\bottomrule
\end{tabular}
\caption{
Final-step KL divergence (bits/byte) under the same target/base-model settings as Table~\ref{tab:ft-kl}, evaluated on another 10,000-text sample from the Pile.
}
\label{tab:ft-kl-other-pile-sample}
\end{table}

\begin{table}[t]
\centering
\scriptsize
\begin{tabular}{llcc}
\toprule
\multirow{2}{*}{Target} & \multirow{2}{*}{Method} & \multicolumn{2}{c}{Base} \\
\cmidrule(lr){3-4}
 &  & random-init & pretrained \\
\midrule

\multirow{2}{*}{Gemma-2B}
& Original & 15.8 & 11.3 \\
& Aggregated-LLD & 13.9 & 9.6 \\
\midrule

\multirow{2}{*}{CodeGemma-2B}
& Original & 10.8 & 9.9 \\
& Aggregated-LLD & 8.4 & 7.4 \\
\bottomrule
\end{tabular}
\caption{
Final-step KL divergence (bits/byte) under the same target/base-model settings as Table~\ref{tab:ft-kl}, evaluated on 10,000 texts sampled from SlimPajama.
}
\label{tab:ft-kl-other-redpajama}
\end{table}

\section{Ablation on Temperature}\label{app:tau-ablation}

\begin{table}[t]
\centering
\scriptsize
\begin{tabular}{lcccc}
\toprule
 Base& $\tau=0.1$ & $\tau=1.0$ & $\tau=10$ & $\tau=\infty$\textsuperscript{*} \\
\midrule
Random-init & 7.31 & {\bf 3.42} & 4.68 & 4.93 \\
Pretrained  & 5.83 & {\bf 2.84} &3.98 & 4.54 \\
\bottomrule
\end{tabular}\\
\vspace{2pt}
\footnotesize \textsuperscript{*} $\tau=\infty$ corresponds to the Pile-original weights.
\caption{KL divergence (bits/byte) to the target model, CodeNanoGPT, under different values of \(\tau\).}
\label{tab:tau-ablation}
\end{table}

The temperature $\tau$ in \eqref{eq:raw-domain-ratio} controls the sharpness of the estimated domain weights. In this appendix, we examine the effect of $\tau \in \{0.1, 1.0, 10, \infty\}$ using randomly initialized or pretrained NanoGPT as the base model and CodeNanoGPT as the target model, where $\tau=\infty$ corresponds to Pile-original weights. As shown in Table~\ref{tab:tau-ablation}, when trained with aggregated-LLD, the KL divergence at the final training step from the target model is minimized at $\tau=1.0$ for both randomly initialized and pretrained NanoGPT.

A low softmax temperature concentrates the domain weights on a few domains, whereas a high temperature makes them closer to Pile-original weights. The difference in mean log-likelihood per token between randomly initialized NanoGPT and CodeNanoGPT ranged from 7.3 (Pile-CC) to 9.1 (DM Mathematics), giving a spread of 1.8. Choosing a temperature on the same scale as this range helps avoid extreme outcomes.

\section{Likelihood-Based Downstream Evaluation}
\label{app:downstream_bpb}
\begin{figure}[t]
    \centering
    \includegraphics[width=1.0\linewidth]{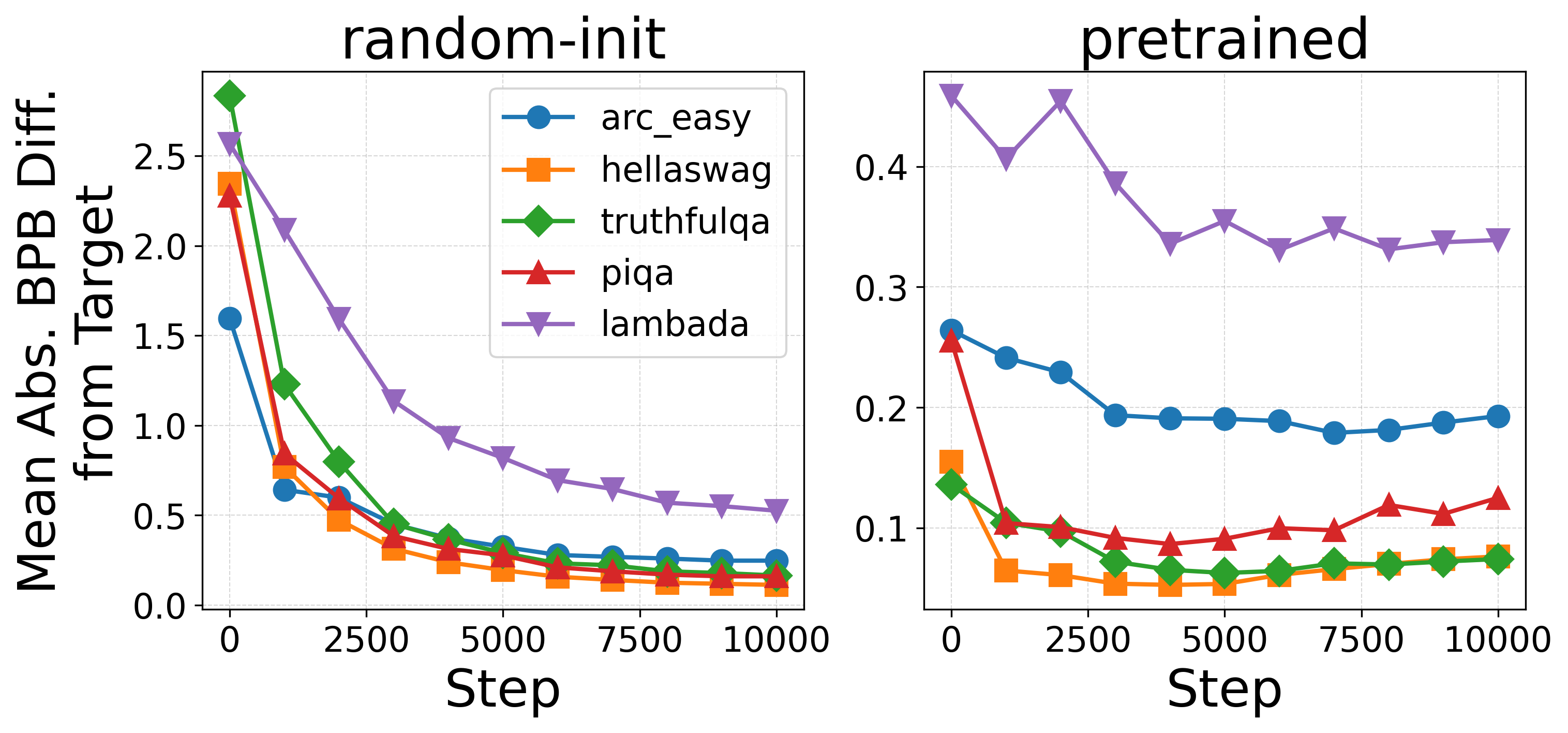}
    \caption{Likelihood-based downstream similarity to the target model.
    For each task, we compute the BPB of the correct completion and report the mean absolute BPB difference between the trained base model and the target model. Lower values indicate that the trained model assigns likelihoods more similarly to the target model.
    The left and right panels show randomly initialized and pretrained NanoGPT, respectively.
    }
    \label{fig:downstream_bpb}
\end{figure}

Accuracy-based downstream evaluation can be noisy or insensitive when models are small and training budgets are limited.
\citet{heineman2025signalnoiseframeworkreducing} argue that likelihood-based metrics such as perplexity or BPB can provide a higher signal-to-noise diagnostic for small-scale model development, and compute BPB on correct continuations by normalizing the negative log-likelihood of the correct answer by its number of UTF-8 bytes.
Following this motivation, we evaluate downstream likelihood similarity using BPB on correct completions.

For each example, we compute the BPB of the correct completion by normalizing its negative log-likelihood by the number of bytes in the completion. We then report the mean absolute BPB difference between the base and target model.
Lower values indicate that the trained model assigns likelihoods to correct completions more similarly to the target model.

For multiple-choice tasks, BPB is computed only on the answer completion, not on the full prompt. For ARC-Easy, HellaSwag, and PIQA, we use the gold option specified by the task label.
For LAMBADA, we treat the target word as the completion.
For TruthfulQA, which may contain multiple true answers in the MC2 setting, we compute BPB for each true answer and average over the true answers.

As shown in Figure~\ref{fig:downstream_bpb}, the mean absolute BPB difference from the target model decreases clearly across most tasks in the randomly initialized setting.
This suggests that the proposed mixture also moves the model closer to the target in downstream likelihood profiles, beyond the evaluation corpus used for estimating approximate KL divergence.
In the pretrained setting, the initial differences are already much smaller, and the trends are more modest and task-dependent.

\section{Aggregation of Domain Weights}\label{app:aggregate}
\subsection{Differences across aggregation methods}

\begin{figure}[t]
    \centering
    \includegraphics[width=1.0\linewidth]{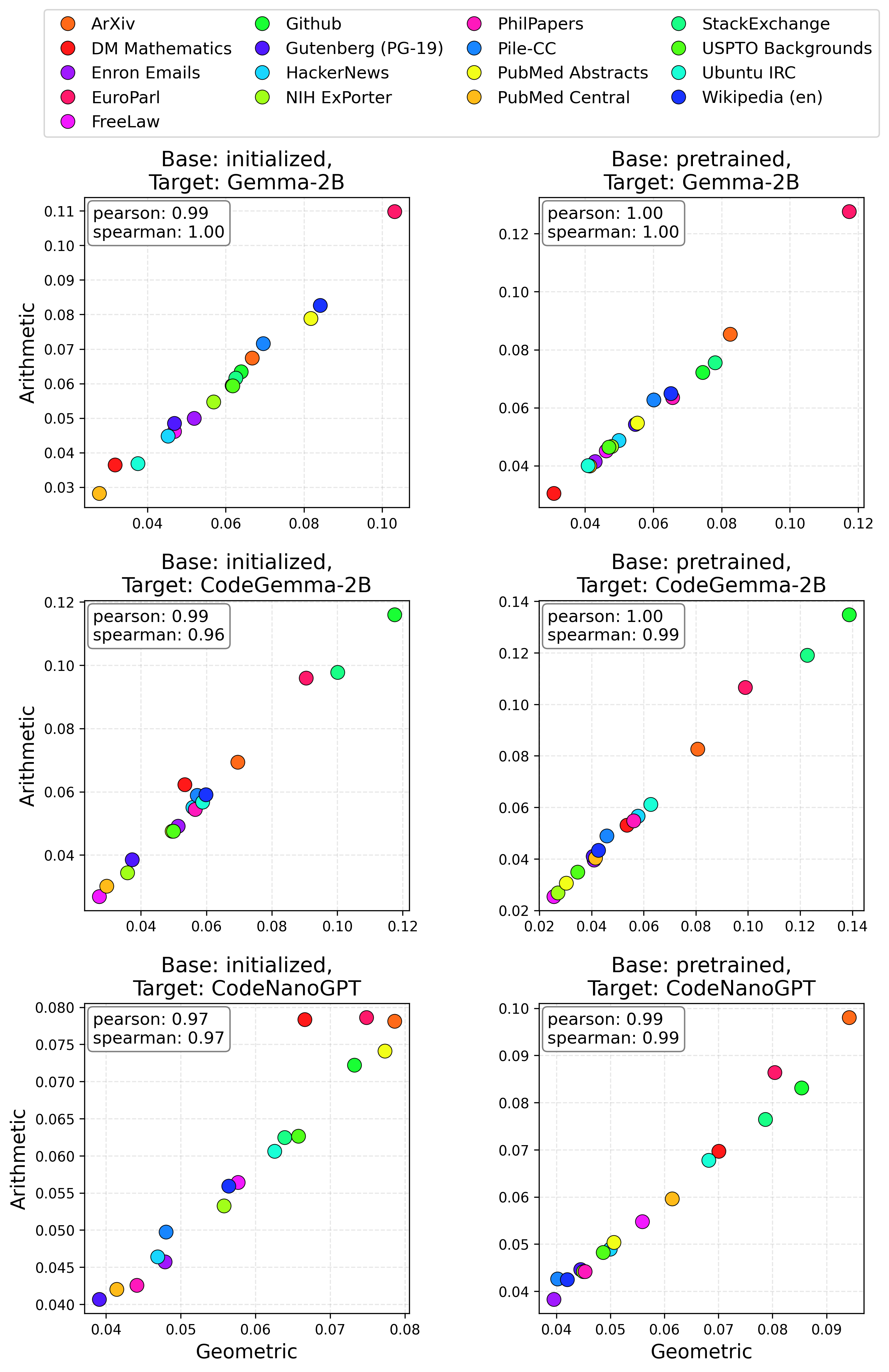}
    \caption{
    Comparison between the domain weights obtained by geometric-mean aggregation (x-axis) and those obtained by arithmetic-mean aggregation (y-axis). Each point corresponds to a domain in the Pile. Scatter plots are shown for each combination of base model (random-init/pretrained) and target model (Gemma-2B, CodeGemma-2B, and CodeNanoGPT).
    }
    \label{fig:pi-geometric-arithmetic}
\end{figure}

In the main text, we obtained $\bm\pi^*$ by taking the geometric mean of $\{\bm\pi^{(t)}\}_{t\in\mathcal T}$, which are estimated by iterative-LLD, as in \eqref{eq:domain-ratio-aggregate}. This aggregation method is based on the optimization problem in \eqref{eq:domain-ratio-aggregate-optimization}. In this appendix, we consider aggregation by the simple arithmetic mean,
\[
\bar\pi_k = \frac{1}{|\mathcal{T}|}\sum_{t\in\mathcal T} \pi_k^{(t)},
\]
and analyze how the resulting domain weights differ depending on the choice of aggregation method.

The target models are the three models considered in the main text: Gemma-2B, CodeGemma-2B, and CodeNanoGPT. The base models are also the two models considered in the main text: random-init NanoGPT and pretrained NanoGPT. Fig.~\ref{fig:pi-geometric-arithmetic} shows, for each base--target model pair, scatter plots comparing the geometric and arithmetic means of $\{\bm\pi^{(t)}\}_{t\in\mathcal T}$ obtained by iterative-LLD. A high correlation is observed between the two methods. This indicates that, regardless of whether geometric-mean aggregation or arithmetic-mean aggregation is used, the relative importance and ranking of the domains are almost unchanged. Therefore, although the geometric-mean aggregation used in this study is theoretically natural from the viewpoint of minimizing the KL divergence in \eqref{eq:domain-ratio-aggregate-optimization}, it is also empirically confirmed to be a stable aggregation method that yields domain weights consistent with those obtained by arithmetic-mean aggregation\footnote{When the variation of $\{\pi_k^{(t)}\}_{t\in\mathcal T}$ is small for each $k$, a Taylor expansion of $\log \pi_k^{(t)}$ around its mean shows that the first-order term vanishes when averaged. This suggests that the geometric mean yields a value close to the arithmetic mean.}.
\end{document}